\documentclass[11pt,a4paper]{article}
\usepackage{times,latexsym}
\usepackage{url}
\usepackage[T1]{fontenc}

\usepackage[acceptedWithA]{tacl2021v1}

\usepackage{xspace,mfirstuc,tabulary}

\newif\iftaclinstructions
\taclinstructionsfalse %
\iftaclinstructions
\renewcommand{\confidential}{}
\renewcommand{\anonsubtext}{(No author info supplied here, for consistency with
TACL-submission anonymization requirements)}
\newcommand{\instr}
\fi

\iftaclpubformat %

\else

\fi

\usepackage{tikz-qtree}
\usepackage{booktabs}       %
\usepackage{amsfonts}       %
\usepackage{subcaption}
\usepackage{todonotes}

\usepackage{xspace}
\usepackage{xparse}
\newcommand{\model}{SiLM}

\NewDocumentCommand{\alm}{o o}{%
  \ensuremath{%
    \mathrm{ALM}%
    \IfValueT{#1}{_{#1}}%
    \IfValueT{#2}{^{#2}}%
  }\xspace
}
\NewDocumentCommand{\tf}{o o}{%
  \ensuremath{%
    \mathrm{MLM}%
    \IfValueT{#1}{_{#1}}%
    \IfValueT{#2}{^{#2}}%
  }\xspace
}
\NewDocumentCommand{\structformer}{o o}{%
  \ensuremath{%
    \mathrm{StructFormer}%
    \IfValueT{#1}{_{#1}}%
    \IfValueT{#2}{^{#2}}%
  }\xspace
}
\NewDocumentCommand{\sfrmr}{o o}{%
  \ensuremath{%
    \mathrm{SF}%
    \IfValueT{#1}{_{#1}}%
    \IfValueT{#2}{^{#2}}%
  }\xspace
}
\NewDocumentCommand{\udgn}{o o}{%
  \ensuremath{%
    \mathrm{UDGN}%
    \IfValueT{#1}{_{#1}}%
    \IfValueT{#2}{^{#2}}%
  }\xspace
}
\NewDocumentCommand{\gpst}{o o}{%
  \ensuremath{%
    \mathrm{GPST}%
    \IfValueT{#1}{_{#1}}%
    \IfValueT{#2}{^{#2}}%
  }\xspace
}
\NewDocumentCommand{\bfgpst}{o o}{%
  \ensuremath{%
    \mathrm{\textbf{GPST}}%
    \IfValueT{#1}{_{#1}}%
    \IfValueT{#2}{^{#2}}%
  }\xspace
}

\title{%
Understanding Syntactic Generalization in \\Structure-inducing Language Models
}

\author{
  David Arps$^\diamond$ 
  \and
  Hassan Sajjad$^\dagger$
  \and
  Laura Kallmeyer$^\diamond$ 
  \\
  $^\diamond$Heinrich-Heine-Universität, Düsseldorf, Germany\\
  $^\dagger$Dalhousie University, Halifax, Canada\\
  $^\diamond$\texttt{first.last@hhu.de}\ \ \ 
  $^\dagger$\texttt{HSajjad@dal.ca}
}

\date{}

\begin{document}
\maketitle

\begin{abstract}
Structure-inducing Language Models (SiLM) are trained on a self-supervised language modeling task, and induce a hierarchical sentence representation as a byproduct when processing an input. 
SiLMs couple strong syntactic generalization behavior with competitive performance on various NLP tasks, but many of their basic properties are yet underexplored. 
In this work, we train three different SiLM architectures from scratch:  Structformer \cite{shen-etal-2021-structformer}, UDGN \cite{shen-etal-2022-unsupervised}, and GPST \cite{hu-etal-2024-generative}.
We train these architectures on both natural language (English, German, and Chinese) corpora and synthetic bracketing expressions. The models are then evaluated 
 with respect to 
(i) properties of the induced syntactic representations %
(ii) performance on grammaticality judgment tasks, and
(iii) training dynamics. %
We find that none of the three architectures dominates across all evaluation metrics. 
However, there are significant differences, in particular with respect to the induced syntactic representations. 
The Generative Pretrained Structured Transformer (GPST; Hu et al. 2024) performs most consistently across evaluation settings, and outperforms the other models on long-distance dependencies in bracketing expressions. 
Furthermore, our study shows that small models trained on large amounts of synthetic data provide a useful testbed for evaluating basic model properties.
\end{abstract}

\section{Introduction}

Linguistic research has 
shown that sentence structure has many hierarchical properties \citep[see, e.g.,][]{Radford:2004,vanValin:2005}. 
While the majority of mainstream large language models process sentences in a strictly sequential fashion,  
over the past years, a number of language modeling architectures have been proposed that integrate self-supervised language modeling tasks with hierarchical sentence structure induction. 
These models have demonstrated competitive performance on NLP tasks, unsupervised parsing, and grammatical generalization benchmarks~\cite{hu-etal-2024-generative,shen-etal-2022-unsupervised,momen-etal-2023-increasing} but many of their basic properties, demonstrating their strengths from learning a hierarchical structure, are yet underexplored.
For instance, \citet{williams-etal-2018-latent} find that \citet{yogatama-etal-2017-learning}'s RL-SPINN model falls back on trivial baseline syntactic structures, and \citet{choi-etal-2018-learning}'s ST-GUMBEL induces structures that vary strongly when the model is trained repeatedly. 
In addition, many of the models are trained on relatively small datasets such as the PennTreebank \cite[1M words,][]{marcus-etal-1993-building} or BLLIP \cite[30M words,][]{charniak-etal-2000-bllip}, which %
leaves the question of their scalability to large datasets unanswered.

This calls for a \textbf{systematic comparison of self-supervised hierarchically biased language models} to understand their strengths and limitations and their potential to becoming a competitive architecture for NLP applications and linguistic research. 
In this work, we aim to fill this gap by conducting a comprehensive study covering aspects of architecture, scalability, self-consistency over several training runs, and syntactic generalizations learned by the models. 
To this end, besides using data from three natural languages, we provide new benchmark data for testing such models, based on formal languages that exhibit non-trivial syntactic structures. 
Furthermore, we choose three substantially different model architectures and compare them with respect to the syntactic structures they induce when being trained either on English data or on synthetic formal language data. 
As a result, this paper contributes a systematic evaluation, benchmark data and evaluation pipeline that inform the design and evaluation of structure-inducing language models.  %
Concretely, %
 we make several contributions. 

\paragraph{Existing Approaches} In Section~\ref{sec:ltlms}, we take stock of the different approaches that have been proposed to combine language modeling with constituency or dependency parsing. 
Out of these, we %
 select three representative model architectures: 
The \textit{Unsupervised Dependency Graph Network} \cite[UDGN]{shen-etal-2022-unsupervised} relies on a bidirectional LSTM parser and constrains self-attention using non-projective dependency graphs.
The \textit{StructFormer} \cite{shen-etal-2021-structformer}, on the other hand, extends a transformer encoder with an unsupervised CNN-based parser that constrains self-attention using projective dependency graphs. 
Both UDGN and StructFormer are trained on masked language modeling, and both induce probabilistic adjacency matrices of a syntactic dependency graph. 
The \textit{Generative Pretrained Structured Transformer} \cite[GPST]{hu-etal-2024-generative} induces a binary constituency tree from an inside-outside autoencoder. 
Span representations from this autoencoder are fed into a generative transformer that generates a sequence of shift-reduce operations together with the text sequence. 

\paragraph{Datasets} In Section~\ref{sec:datasets}, we describe the English, German, and Chinese datasets that we use for training these models. 
Additionally, we test the structural learning and generalization capabilities of these \model\ architectures, and encoder and decoder transformer baselines. 
We use Dyck languages -- well-formed bracketings of $k$ different bracket types -- which lack structural ambiguities and for which the underlying structure is known \cite{chomsky-schuetzenberger-1959-algebraic,hewitt-etal-2020-rnns}.   
To make up for the lack of ambiguities in Dyck-$k$ languages of $k$ bracket types, we introduce the Dyck-$u$ language that mimics subject-verb agreement in natural languages, and includes an unspecified bracket type. 
To evaluate performance of both natural and formal language models, we use minimal pair benchmarks which link grammaticality and sequence probability without the need for finetuning \cite{warstadt-etal-2020-blimp}: For a grammatical sentence and an ungrammatical counterpart, the grammatical sentence should be treated as more likely. 
We contribute such a minimal pair benchmark for Dyck languages by perturbing bracketing strings in closely defined ways, and with respect to different dependency lengths. 
We argue that this is an efficient way to phrase recognition in formal languages via sequence probabilities. 
This is necessary for evaluating self-supervised models on formal languages in a similar way as these models would be used with natural languages.

\paragraph{Training and Evaluation} 
We present training details in Section~\ref{sec:training}. Section~\ref{sec:evaluation} evaluates the models with respect to several properties of the induced syntactic structures, and their syntactic generalizations. 
We find that, while the \model s are close on some aspects of performance, the formal languages provide a useful testbed for estimating the capabilities of \model s. %
 \gpst
 is most consistent across different evaluation aspects, and outperforms both transformers and other \model s on minimal pair evaluations. 
\textbf{All \model\ architectures show significant variation in induced syntactic representations when retraining identical architectures on the same data}.\footnote{Code and models available at \url{https://github.com/davidarps/silm}}
We conclude the paper with a summary of our findings (Sec.~\ref{sec:conclusion}), and a discussion of open issues (Sec.~\ref{sec:openissues}).

\section{Structure-Inducing Language Models}\label{sec:ltlms}

We define Structure-inducing Language Models (\model s) as neural architectures that are trained on a self-supervised language modeling task, and produce some syntactic representation $t_x$ as a byproduct when processing an input sequence $x$.
The exact nature of $t_x$ may differ between \model\ architectures, but all \model s
 share the property that $t_x$ must be learned in an unsupervised way: 
  No annotated syntactic trees are available during training. 
The nature of the self-supervised language modeling task may vary. 

We focus on masked language modeling tasks (\sfrmr and \udgn), and autoregressive language modeling preceded by span encodings (\gpst). 
The most crucial difference for building $t_x$ is that in masked language modeling, bidirectional context is available for constructing token predictions and $t_x$; while in autoregressive language modeling, both the token predictions and syntactic representations are built incrementally. 
The nature of the syntactic inductive bias shaping $t_x$, as well as the neural architecture that builds $t_x$ and the neural LM architecture may also vary. 
In the following sections, we provide a brief overview over existing approaches, 
and describe in more detail the models we explore. 

\subsection{Related Work}

A wide range of work since the 1990s connects unsupervised parsing and language modeling tasks using neural architectures \cite{sun-etal-1993-neural,chen-1995-bayesian,hihi-bengio-1995-hierarchical,schmidhuber-1991-neural}.
Existing work on the connection of unsupervised parsing and language modeling can be partitioned into several broad categories depending on %
the central backbone of the neural architectures used. 
These include approaches based on Transformers and self-attention, which utilize the fact that attention heads naturally learn to track syntactic relations \cite{shen-etal-2021-structformer,he-etal-2024-language,li-etal-2020-heads,li-lu-2023-contextual,wang-etal-2019-tree,zeng-xiong-2022-unsupervised}.

RNN-based approaches are often augmented with Tree-LSTMs, stacks, and other data structures related to transition-based parsing \cite{bowman-etal-2016-fast,choi-etal-2018-learning,grefenstette-etal-2015-learning,htut-etal-2018-grammar,jacob-etal-2018-learning,kim-etal-2019-unsupervised,li-etal-2019-dependency,shen-etal-2018-neural,yogatama-etal-2018-memory}.
\citet{shen-etal-2022-unsupervised} combines self-attention and LSTMs into a single architecture as
described in Sec.~\ref{sec:model-description}. 

Finally, several methods are inspired by parsing algorithms such as inside and outside span representations and chart parsing \cite{drozdov-etal-2019-unsupervised-latent,drozdov-etal-2020-unsupervised,hu-etal-2021-r2d2,hu-etal-2022-fast,hu-etal-2024-generative,hu-etal-2024-augmenting}. 
\citet{williams-etal-2018-latent} investigate SPINN \cite{yogatama-etal-2017-learning} and ST-Gumbel \cite{choi-etal-2018-learning} architectures. They find that while positive results on NLI can be replicated, the models induce trivially left-branching trees\footnote{left-branching (right-branching) trees are trees where all prefixes (suffixes) of a sentence form constituents} (SPINN); or trees induced from models trained on random seeds have a relatively low similarity in the case of ST-Gumbel.

\citet{ishii-miyao-2023-tree} analyze the interaction of algorithm, dataset and branching bias of several models: PRPN \citep{shen-etal-2018-neural}, DIORA \citep{drozdov-etal-2019-unsupervised-latent} and URNNG \citep{kim-etal-2019-unsupervised}. They train models on English and Japanese, as well as perturbations of the datasets that minimize branching bias. They find that while DIORA has little branching bias, URNNG and PRPN exhibit a right-branching bias. 
With some exceptions \cite{kann-etal-2019-neural,han-etal-2019-multilingual,yang-etal-2023-unsupervised,ishii-miyao-2023-tree,li-etal-2020-empirical,li-lu-2023-contextual,kim-etal-2019-compound,jin-etal-2021-character-based}, the majority of \model s have been only trained and evaluated on English.
However, some works include evaluation on formal languages %
and synthetic data \cite{chowdhury-caragea-2023-beam,grefenstette-etal-2015-learning,ishii-miyao-2023-tree,jacob-etal-2018-learning,li-lu-2023-contextual}. 
\citet{hewitt-etal-2020-rnns} have shown theoretical bounds and practical capabilities of recurrent models for learning Dyck languages with $k$ bracket types and maximum bracketing depth $m$. 
Their results show that LSTMs are capable of generating closing brackets correctly, given enough training data. 
More information about related work can be found in Appendix~\ref{app:relatedwork}.

\subsection{Models for Empirical Comparison}\label{sec:model-description}

In this work, we evaluate three \model\ architectures from the categories discussed above: 
StructFormer \cite{shen-etal-2021-structformer} uses a transformer encoder and a CNN parser module that introduces a syntactic inductive bias into the self-attention mechanism. %
UDGN \cite{shen-etal-2022-unsupervised} combines a bi-LSTM with syntactic multi-head attention. 
It thus represents the recurrency-based approaches in the literature. 
GPST \cite{hu-etal-2024-generative} combines a bidirectional inside-outside autoencoder and a generative transformer with a shift-reduce component, all of which are frequently used in the literature. However, the transformer backbone and the relatively large model scale in \citet{hu-etal-2024-generative} suggests favorable scaling capabilities.%

\paragraph{StructFormer} integrates a Transformer encoder and an unsupervised CNN parser.
We use the implementation from \citet{momen-etal-2023-increasing}, which puts the unsupervised parser in the middle layers of the transformer backbone. 
This is opposed to \citet{shen-etal-2021-structformer}'s version, where the parser is used at the embedding layer. The motivation is that in this way, sequential and hierarchical sentence structure are mixed in a more intuitive way. 

The architecture consists of a number of $l_{front}$ transformer encoder layers that contextualize the input sequence $x$ in a strictly sequential fashion. 
After this, a convolutional network with two feed-forward outputs predicts 
(i) for each pair of tokens $x_i,x_{i+1}$ the distance $\delta_{i}$ in a syntactic tree $t_x$, and  %
(ii) for each token $x_i$ the height $\tau_i$ of this token in a syntactic tree.
The values of $\delta$ and $\tau$ are ranked, and determine the probabilities for dependency relations between tokens. %
This process is described in detail in \cite[Sec. 4.2]{shen-etal-2021-structformer}. 
Concretely, the quadratic heads matrix $H \in \mathbb{R}^{n \times n}$, contains the probability that $x_i$ is a dependent of $j$ at $H[i,j]$. $n$ is the length of the input sequence including BOS and EOS tokens. 
$H$ constrains the multi-head self attention in the remaining $l_{back}$ transformer layers: 

To accommodate the empirical finding that different attention heads track different syntactic relations, 
a constraining set of independent probabilities $q_{i,j}$ per token pair $x_i,x_j$ and per attention head is learned to determine how much the respective head can attend from $x_i$ to $x_j$. 
These probabilities $q_{i,j}$ are then multiplied with the output from self-attention, in order to obtain the final output of each attention head \cite[Eq. 17]{shen-etal-2021-structformer}. 
To eliminate the possibility that a token is a dependent of itself, all values $H[k,k]$ are set to $0$ for $1\le k \le n$. 
The outputs of the last of the $l_{back}$ transformer layers are then projected on the vocabulary using a language modeling head. 

We optimize the model using masked language modeling, such that a single backpropagation pass through the parser module learns an unsupervised dependency tree. 
$H$ can be interpreted as a constituency tree (using only $\delta$), or as a dependency graph (by converting the constituency tree and using the heights $\tau$ for directionality). 
However, since we are primarily interested in the syntactic inductive bias \textit{used} by the model, we treat $H$ as a weighted adjacency matrix for a dependency graph. 
During evaluation, $H$ is converted to a discrete dependency graph $D$ by choosing for each token $x_i$ the single head $x_j$ that %
maximizes the dependency head score for $x_i$ in $H$, i.e., $j=\arg\max_k H_{i,k}$
$D$ is possibly disconnected, and not rooted; with these properties, it follows the intuition of processing the local syntactic context relevant for token-level predictions. 

\paragraph{UDGN}
\citet{shen-etal-2022-unsupervised} shares the motivation with \structformer that self-attention is useful for tracking dependency relations, and that these relations can be learned in an unsupervised way. 
Specifically, an unsupervised biLSTM-based parser predicts a dependency graph. 
The soft adjacency matrix of that graph $H$ is used as input to a \textit{Dependency Graph Network} (DGN) in an unsupervised way. 
The DGN consists of several layers of gated multi-head attention and is optimized via masked language modeling. 
The DGN's gating mechanism is optimized to select an appropriate head for each pair of tokens. 
Similar to the StructFormer, instead of decoding $H$ to obtain a dependency tree, we directly evaluate $H$. 

\paragraph{GPST}
\citet{hu-etal-2024-generative} follows a completely different approach in that it induces constituency instead of dependency trees, uses a combination of self-supervised loss functions, and adds a generative component. 
The GPST architecture consists of several modules: 
A sparse inside-outside autoencoder $\mathbf{ae}$ predicts span representations in a binary constituent tree, based on a two-step process: 
Using a transformer-based parser as a pruning heuristic \cite{hu-etal-2024-augmenting}, a bottom-up pass through the possible binary constituent trees for $x$ computes inside representations $\mathbf{i}_{i,j}$. 
A top-down pass, again using transformer encoders, computes outside representations $\mathbf{o}_{i,j}$. 
These can be viewed as representations of the context of $x_{i:j}$, that is, all tokens that are not in $x_{i:j}$. 
The outside representations $\mathbf{o}_i$ of any token are then used to predict a probability distribution over the vocabulary, which is optimized as a bidirectional language modeling loss $\mathcal{L}_{ae}$. 

The second module is an autoregressive transformer $\mathit{TF}$ that processes the sentence incrementally. 
$\mathit{TF}$ takes the inside scores $\mathbf{i}$ as input embeddings, which serve as compact representations for any possible span in the parse tree. 
$\mathit{TF}$ consists of two parts, a $\mathit{TF}_{action}$ transformer predicts the actions of a shift-reduce parser that generate the tree induced by the autoencoder \textbf{ae}, and an autoregressive $\mathit{TF}_{lm}$ transformer predicts the next token.  
The first step models a shift-reduce pass over the sequence, building a binary constituent tree during processing the sentence using discrete actions and a stack of past representations. 
Concretely, using $l_{action}$ transformer layers $\mathit{TF}_{action}$, the model decides whether to \texttt{GEN}erate the next token and shift it onto a stack $\Gamma$ of processed tokens, or whether to \texttt{COMP}ose the top two constituents from $\Gamma$. 
For the \texttt{GEN}eration step, an autoregressive Transformer $\mathit{TF}_{lm}$ is called. $\mathit{TF}_{lm}$ takes the output of the final layer of $\mathit{TF}_{action}$ as input, and generates the next token via a standard transformer decoder architecture and language modeling loss. 
$\Gamma_1= \mathbf{x}_{r:s}, \Gamma_0 = \mathbf{x}_{s:t}$ are removed from $\Gamma$, and a new constituent $\mathbf{x}_{r:t}$ is added at the top of the stack. 

The generative transformer models are optimized using a complex loss function, consisting of an autoregressive language modeling loss $\mathcal{L}_{ntp}$, which consists of cross-entropy losses for the token predictions $\mathcal{L}_{ntp}$ as well as stack action predictions ($\mathcal{L}_{ar} = \mathcal{L}_{ntp} + \mathcal{L}_{action}$). 
To mitigate a left-branching bias in induced trees, the minimization of $\mathcal{L}_{ar}$ via backpropagation partially ignores the parameters of the autoencoder. 
The full autoencoder is optimized using an autoencoding loss for inside-outside token representations ($\mathcal{L}_{ae}$), as well
as unsupervised loss functions that ensure balancing of the induced trees and maximize the likelihood of the pruning for span computations.

\section{Datasets}\label{sec:datasets}

We pretrain \model s on different kinds of data (in three natural languages, and on synthetic formal language data). Each training dataset is around 100M tokens; evaluation sets are around 1M tokens for natural languages and 100K tokens per formal language.

\subsection{English}
We use a de-duplicated and cleaned version of the BabyLM 2023 and 2024 data \cite{warstadt-etal-2023-findings,hu-etal-2024-findings}. 
We randomly sample a held-out test set consisting of roughly $1\%$ of the training data size. 
We chose this dataset size for a variety of reasons: 
Models trained on this amount of data have been shown to deliver good performance on a variety of tasks \cite{warstadt-etal-2023-findings}, yet are small enough to be both practical to train, and useful for psycholinguistic research \cite{wilcox-etal-2024-bigger}. 
Most existing \model s have been trained on datasets that are 1-2 orders of magnitude smaller (Sec.~\ref{sec:ltlms}).
We compare the induced $t_x$ to dependency trees parsed with Spacy \cite{Honnibal-etal-2020-spacy}, and constituency trees parsed with SuPar \cite{zhang-etal-2020-fast}.

\subsection{German}
The German pretraining dataset is sampled from three sources, in a similar fashion as the English dataset. 
As a basis, we take the German BabyLM dataset of Child-directed speech from \cite{bunzeck-etal-2025-construction} (16.5M words).
To this, we add roughly 16.9M words of books from various domains (from Project Gutenberg), and 
67.5M words from OpenSubtitles \citep{lison-tiedemann-2016-opensubtitles2016}. 
we compare the induced $t_x$ to dependency trees parsed with Spacy \cite{Honnibal-etal-2020-spacy}, and constituency trees that are obtained from the dependency trees via deterministic conversion: Every subtree in the dependency tree is also a subtree in the constituency tree, where heads and dependents are siblings in the resulting constituency tree. For non-projective dependencies, the discontinuous dependent is attached at a higher internal node to remove the discontinuity.

\subsection{Chinese}
We use the Chinese portion of the BabyBabelLM dataset \citep{jumelet-etal-2025-babybabellm} for pretraining. 
After deduplication, this dataset is 
around 100M tokens in size. Of those, 78M tokens are transcribed (child-appropriate) speech, 12M tokens are books, 9.5M tokens from educational material, and 0.5M tokens child-appropriate wikis.

\begin{figure}\centering
  \begin{tikzpicture}[scale=.8]
    \tikzset{level distance=17pt}
    \tikzset{frontier/.style={distance from root=51pt}}
    \tikzset{edge from parent/.style= {draw,edge from parent path={(\tikzparentnode.south) -| (\tikzchildnode)}}}
    \Tree [  $(_{23}$ [ $(_4$ [ $(_{40}$ $)_{40}$ ] $)_4$ ] [ $(_{51}$ $)_{51}$ ] $)_{23}$ ]
    
  \end{tikzpicture}

  \begin{tikzpicture}
    \tikzset{level distance=17pt}
    \tikzset{frontier/.style={distance from root=51pt}}
    \tikzset{edge from parent/.style= {draw,edge from parent path={(\tikzparentnode.south) -| (\tikzchildnode)}}}
    \Tree[ $(_{u}$ [ $(_{2}$ [ $(_{u}$  $)_{2}$ ] $)_{u}$ ] [ $(_{1}$  $)_{1}$ ]  $)_{1}$ ]
  \end{tikzpicture}
  \caption{Examples from the Dyck-$64$ language (top), and the Dyck-$u$ language (bottom).}
  \label{fig:dyck-example}
\end{figure}
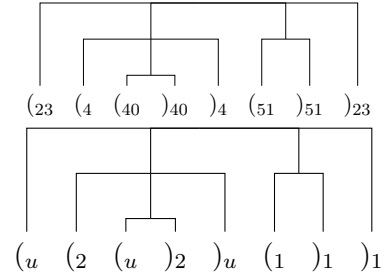

\subsection{Formal languages}
\paragraph{Dyck-$k$} We also train \model s on data generated from formal grammars. 
This data has one crucial difference from natural language %
 data: the distribution from which the data is generated, as well as the true syntactic structure underlying the data, is fully known. 
Furthermore, Dyck languages isolate the dependencies between words into prototypical structures. 
Consider the English sentence \textit{The trees$_{pl}$ that he$_{sg}$ planted$_{sg}$ have$_{pl}$ grown.} 
The center-embedded relative clause features singular number agreement between subject and verb, while the subject and verb of the main clause are plural. 
These (possibly long-distance) dependencies could be represented in a bracketing structure such as $(_{1}\,(_{2} \,)_{2} \,)_{1}$ \citep{karlsson-2007-constraints,hewitt-etal-2020-rnns}.
This parallel between center embedding in natural languages and Dyck languages motivates us to test to which extent \model s are able to learn these kinds of structures in a more isolated and closely controlled environment with a more closely defined vocabulary, and clearly defined syntactic structures.

Concretely, we train models on data from four formal bracketing languages: 
First, we train different models on Dyck-$k$ languages of well-nested bracketings with $k$ bracket types \cite{chomsky-schuetzenberger-1959-algebraic,hewitt-etal-2020-rnns}. 
We use $k\in \{1,2,64\}$ bracket types and maximum bracketing depth up to 7. 
The upper tree in Figure~\ref{fig:dyck-example}, taken from the Dyck-$64$ language, has for instance a maximum bracketing depth of 3 because the deepest nesting (brackets of type 40) is 3 levels deep. 
For the Dyck languages, 
sequences in the train split are up to 96 tokens long.
We evaluate on a validation split of the same maximum length, and on length generalization splits that are up to 192 tokens long. 
We use the 
implementation 
of \citet{hewitt-etal-2020-rnns} to generate Dyck-$k$ data.

\paragraph{Dyck-$u$} We also define the Dyck-$u$ language, which follows well-nested bracketings but where bracket types can be unspecified. 
This language is similar to Dyck-2, with one crucial difference: Each bracket exists in a specified and unspecified format. 
An example from Dyck-$u$ is displayed at the bottom of Figure~\ref{fig:dyck-example}. 
In this language, three types of brackets exist: $1, 2$, and $u$. 
For $1$ and $2$, the same criteria for well-nested bracketings exist as for Dyck-$k$ languages. 
However, a sequence is also in the Dyck-$u$ language if $(_1$ or $(_2$ are closed by a $)_u$. Vice versa, $(_u$ can be closed by either $)_u, )_1$, or $)_2$. 
Specified ($1,2$) or unspecified ($u$) brackets are chosen at uniform probabilities. 
Dyck-$u$ is a simplification of agreement in natural languages. 
For instance, both nouns and verbs can share a form between singular and plural (\textit{The fish$_{u}$ swims$_{sg}$} vs.~\textit{The fish$_{u}$ swim$_{pl}$} and \textit{The children$_{pl}$ arrived$_{u}$} vs.~\textit{The child$_{sg}$ arrived$_{u}$}). 
Train, validation and generalization splits are generated analogously to the Dyck-$k$ languages.

\paragraph{Syntactic Annotations} 
Gold dependency structures for the formal bracketing language are built by putting undirected edges between each pair of opening and closing bracket. 
The resulting graphs are not connected, but projective (i.e., without crossing edges). Both \sfrmr and \udgn models are theoretically capable of inducing them. 
Gold constituency trees are created in the data generation.

\subsection{Minimal pairs}
To evaluate grammatical generalization, we use minimal pair settings that link sentence-level perplexity to grammaticality. 
For the English models, we use the Benchmark of Linguistic Minimal Pairs \cite[BLiMP,][]{warstadt-etal-2020-blimp}. 
The task is that, without any fine-tuning, a model should assign lower perplexity (i.e., higher likelihood) to a grammatical sentence than to its ungrammatical counterpart. 
In BLiMP, minimal pairs are constructed for 12 categories of closely-controlled phenomena such as subject-verb agreement (\textit{Angela likes Connie} vs. \textit{Angela like Connie}). 
For masked models, we obtain perplexity scores using left-to-right subword masking, as proposed by \citet{kauf-ivanova-2023-better} and \citet[Eq.~3]{arps-etal-2024-multilingual}.
For minimal pair evaluation in German, we use the German portions of both CLAMS \cite[Cross-Linguistic Assessment of Models on Syntax,][]{mueller-etal-2020-cross} and multilingual BLiMP \citep[mBLIMP,][]{jumelet-etal-2025-multiblimp} with 9 categories total.
For Chinese, we use the ZhoBLiMP \citep[13 categories]{liu-etal-2024-zhoblimp}.

\begin{table}\centering\scalebox{.75}{
  \begin{tabular}{ll}
\toprule
positive sample & $(_{23}\ (_4\ (_{40}\ )_{40}\ )_4\ (_{51}\ )_{51}\ )_{23}$\\ \midrule
\multicolumn{2}{l}{corresponding negative samples:} \\ %
Method & Result \\
\texttt{bracketswap} & $(_{23}\ (_4\ \mathbf{)_{40}\ (_{40}}\ )_4\ (_{51}\ )_{51}\ )_{23}$ \\
\texttt{randomswap} & $(_{23}\ (_4\ (_{40}\ \mathbf{(_{51}}\ )_4\ \mathbf{)_{40}}\ )_{51}\ )_{23}$ \\
\texttt{typemismatch} & $(_{23}\ (_4\ (_{40}\ )_{40}\ \mathbf{)_5}\ (_{51}\ )_{51}\ )_{23}$ \\
\bottomrule
\end{tabular}}
\caption{Negative samples for minimal pairs. The positive sample is the Dyck-64 string in Figure~\ref{fig:dyck-example}.}
\label{tab:dyck-mp-examples}
\end{table}

For formal languages, we generate a similar benchmark by perturbing well-formed bracketings in three different ways. 
Concretely, we create negative samples for minimal pairs by swapping or replacing parts of the input (Table~\ref{tab:dyck-mp-examples}) in three different ways. 
For \texttt{bracketswap}, matching opening and closing brackets are swapped. 
For \texttt{randomswap}, two randomly sampled tokens are swapped (without the requirement that they are matching brackets), 
and for \texttt{typemismatch}, the type of a single opening or closing bracket is changed. 
Different subtasks are created by swapping brackets with different distances - in this concrete example, a \texttt{bracketswap} for type $40$ creates a local change; whereas swapping brackets of type $23$ would test for more long-range dependencies. 
All negative samples are tested for ungrammaticality.

\section{Pre-Training}
\label{sec:training}
\paragraph{Models}
For each language, we compare performance to a transformer encoder baseline in the style of RoBERTa \cite{liu-etal-2019-roberta}, and a transformer decoder baseline in the style of GPT-2 \cite{radford-etal-2019-language}. 
According to their pre-training objective, these are abbreviated as \tf (masked LM) for the encoder, and \alm (autoregressive LM) for the decoder baseline. 
For each \model s architecture, we train three models with identical training data and hyperparameters, but different random model parameter initializations and random seeds for training data loading. 
For formal language settings, we target each model and the baseline to have around 2M trainable parameters. 
We argue that this is sufficient because the vocabulary of these languages is small. 
Each natural language model has approximately 15M parameters. 
Model dimensions and hyperparameters are listed in App.~\ref{app:modeldims}.
All natural language models use monolingual BPE tokenizers with a vocabulary size of 10000. 
The tokenizers are trained on the respective training data splits, and follow the implementation of RoBERTa \citep{liu-etal-2019-roberta}.

\paragraph{Experimental setup}
In the first step, we empirically determine the hyperparameters by training on the Dyck-$u$ language, and English. 
Then, we train three model instances per architecture and dataset, using the hyperparameters listed in Appendix \ref{app:modeldims}. 
Model instances are trained on a single A100 card each, for 48 hours. 
After this, for the natural language models, we compare the three models from each architecture in terms of test set perplexity,\footnote{For UDGN and StructFormer, we use masked LM pseudo-perplexity as defined by \cite{salazar-etal-2020-masked}} and train the best model for 500K steps in total. 
We assume, for the relatively small formal language models, that 48 hours of training is a sufficient training time. 
For the natural language models, this setup allows to factor in the training efficiency of the different model architectures.
Generally, both training and validation loss are either stable after this period of time (in case of the formal language models), 
or still moderately decrease (for natural language models).

\section{Evaluation}\label{sec:evaluation}

We evaluate the models with respect to the following properties. 
With respect to the induced representations $t_x$, we ask 
\textit{How similar are induced representations $t_x$ between models of the same architecture trained on the same data?} ($t_x$-consistency); 
\textit{Are induced representations similar to trivial baseline representations?} ($t_x$-triviality);
\textit{Over the course of training, are $t_x$ converging to stable representations, or are they jumpy?} ($t_x$-learning-evolution);
\textit{Are induced representations similar to gold representations?} ($t_x$-annotation-similarity).

With respect to model performance on Dyck languages, we evaluate how well the models generalize to sequences of lengths not seen in training. We evaluate the natural language models on (pseudo-)perplexity and minimal pairs, 
and create a similar benchmark for the formal languages that evaluates the models capability to isolate model behavior towards grammaticality, sequence likelihood, and long-distance dependencies. 

In comparing training speed, we first evaluate the intrinsic qualities of the induced representations $t_x$. 
For natural language models, we observe that for multi-token words, the models show a strong trend to select other tokens from the same word as heads. 
For instance, the word \textit{meanwhile} is tokenized as \textit{Ġmean while}, and the row for \textit{Ġmean} in $H$ puts a large amount of probability mass on \textit{while}.
To control for this behavior, we find the heads for multi-token words by excluding the head to be in the same word, and summing over the head distributions for all tokens in the word. 

\begin{figure}\centering
    \includegraphics[clip, trim=0.0cm 0.0cm 0.0cm .0cm, height=2.9cm]{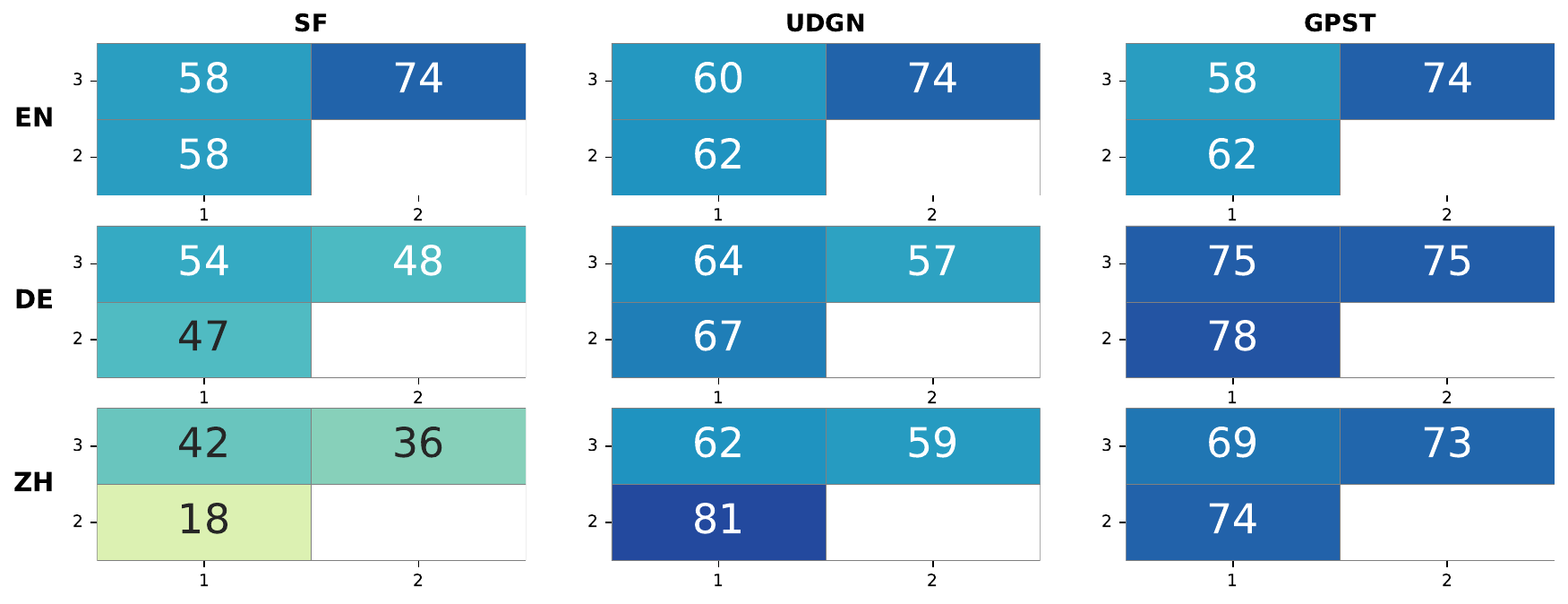}
    \includegraphics[clip, trim=0.0cm 0.2cm 0.0cm .0cm, height=3.5cm]{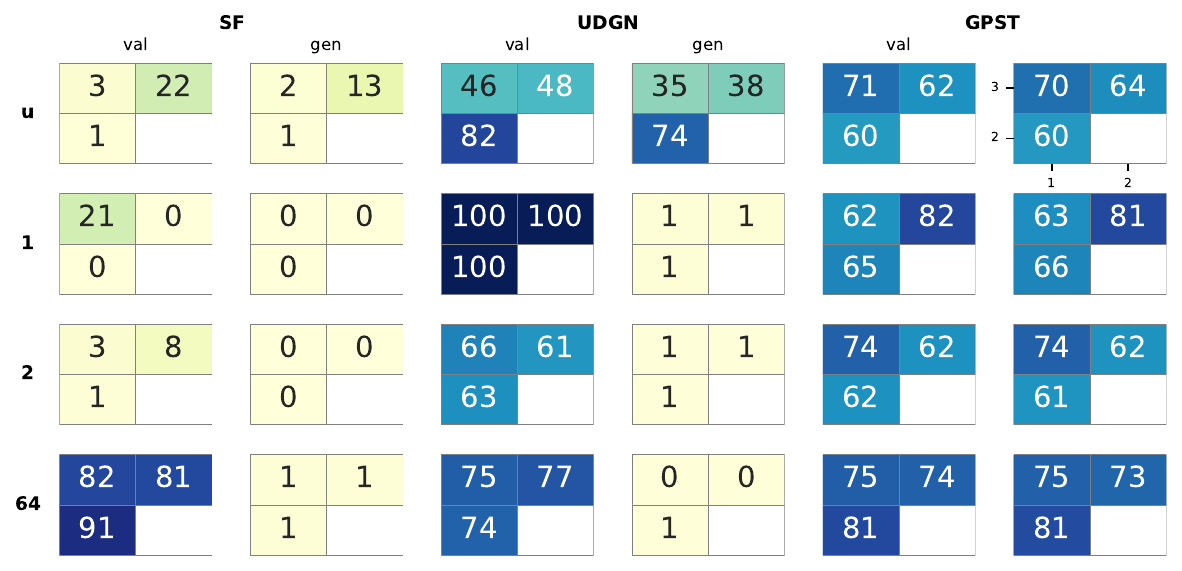}
    \caption{$t_x$-consistency for natural languages (top) and Dyck languages (bottom), measured in UAS for \sfrmr and \udgn, and F score for \gpst.}
\label{fig:t-consistency}
\end{figure}

\paragraph{$t_x$-consistency} 
Figure~\ref{fig:t-consistency} shows that there is significant variation between the syntactic structures induced by different natural language model instances of the same architecture. 
\sfrmr and \udgn are both evaluated via UAS (unlabeled attachment score; the ratio of words for which the correct head is selected), and for both English models, the $t_x$-consistency falls in a very similar range of around 15 points. 
For German and Chinese \sfrmr, the $t_x$-constituency is lower than for English.
Chinese \sfrmr and \udgn models have the most variance wrt. $t_x$-consistency, reaching from 18-42 UAS for \sfrmr and from 59-81 for \udgn. 
The F-scores for $t_x$-consistency of \gpst models indicate an opposite trend (but the difference in metric does not allow a direct comparison):  For all languages, F scores are relatively high (between 58 and 78), but the $t_x$-consistency of German is highest (followed by Chinese, and then English).

For the formal languages, however, the picture is different: 
Only the $t_x$-consistency of \gpst models is consistently high on both validation and generalization sets. 
UAS for \sfrmr and \udgn are generally very low on generalization sets, indicating that almost no generalization to longer sequence lengths happens for either architecture. 
Investigation on the heads matrices $H$ in \sfrmr formal language models shows that the main reason for this is that the trained \sfrmr models put similar head probabilities on many possible head tokens, yielding almost uniform distributions in $H$. This behavior is explored further in  App.~\ref{app:sf-uniformheads}.
The only exception is one pair of \udgn models trained on the Dyck-$u$ language.
On validation sets, UAS for \udgn are generally higher than for \sfrmr, which show very low similarities except for \sfrmr trained on Dyck-$64$. 
For the UDGN models trained on Dyck-$u$, we see that \udgn[1,u] and \udgn[2,u] induce relatively similar trees (.82), while the UAS between other model pairs are lower.
To ensure a fair comparison with respect to the amount of training, we compare the checkpoints taken after the maximum number of training steps that the two instances have trained for.

\begin{table}\centering\scalebox{.75}{
\begin{tabular}{llrrrr}
\toprule 
UAS & & first & last & prev & next \\
\midrule
SF   & en & 2 & 0 & 39 & \textbf{47} \\
UDGN & en & 7 & 6 & 14 & \textbf{47} \\
SF   & de & 27 & 14 & 31 & 17 \\
UDGN & de & 8 & 6 & 31 & 17 \\
SF   & zh & 15 & 12 & 41 & 2 \\
UDGN & zh & 5 & 6 & 48 & 25 \\
\bottomrule
\end{tabular}}

\scalebox{.75}{
\begin{tabular}{llrr}
\toprule
F & & left-branch & right-branch \\
\midrule
GPST & en & 16 & 21 \\
GPST & de & 30 & 20 \\
GPST & zh & 17 & 20 \\
\bottomrule
\end{tabular} }
\caption{$t_x$ induction baselines on the natural language test sets. 
    first and last are trivial trees where the head of every word is the BOS or EOS token. 
    prev and next are trivial trees where the head is always the previous or next word.} %
    \label{tab:tree-triviality-nl}
\end{table}

\paragraph{$t_x$-triviality} 
Table\ \ref{tab:tree-triviality-nl} shows that natural language \sfrmr and \udgn models trained for 500K steps induce $t_x$ that are clearly different from trivial baseline trees. 
With the exception of German \sfrmr, 
BOS and EOS tokens are rarely selected as heads. 
Both model architectures have a tendency to select adjacent words as the head. 
For English, the next words is preferred over the previous word for both models, while for 
 German and Chinese, this trend is reverted to favoring the previous word as head. 

On the formal languages, \udgn induces $t_x$ with low similarities to trivial baseline trees in all settings except the Dyck-$1$ language, where edges always lead to either the BOS or EOS token (Appendix~\ref{app:moreresults}). 
This suggests that in the case of Dyck-$1$, the small vocabulary prevents the model from inducing nontrivial representations. 
For \gpst models on all languages, we observe that the trees have low similarity with left- or right-branching trees (the highest similarity being an F of 30 for German \gpst and left-branching trees). 

\begin{table*}\centering
  \scalebox{.75}{
\begin{tabular}{ll|rr|rr||rrrrrr}
    \toprule
    & & \multicolumn{2}{c}{\sfrmr} & \multicolumn{2}{c}{\udgn} & \multicolumn{6}{c}{\gpst} \\
    & model & val & gen & val & gen & val P & gen P & val R & gen R & val F & gen F \\
    \midrule
      & 1 & 8.1 & 9.7 & 49.5 & 46.2 & 31.9 & 30.2 & 59.3 & 56.4 & 41.5 & 39.3 \\
    u & 2 & 3.7 & 2.3 & 47.6 & 43.2 & 36.5 & \textbf{34.9 }& 67.9 & \textbf{65.4 }& 47.5 & 45.5 \\
      & 3 & 5.2 & 3.7 & \textbf{78.4} & \textbf{58.9} & \textbf{36.7} & 34.7 & \textbf{68.3} & 64.9 & 47.7 & 45.2 \\\hline
      & 1 & 6.3 & 4.6 & 7.6 & 5.4 & \textbf{27.0} & \textbf{26.6} & \textbf{50.3} & \textbf{49.7} & 35.2 & 34.6 \\
    1 & 2 & 3.4 & 2.6 & 7.6 & 5.4 & 25.8 & 25.6 & 48.2 & 48.2 & 33.6 & 33.5 \\
      & 3 & 4.4 & 3.1 & 7.6 & 5.4 & 24.7 & 24.6 & 46.3 & 46.3 & 32.2 & 32.1 \\\hline
      & 1 & 6.5 & 5.3 &\textbf{ 67.0} & \textbf{61.5} & \textbf{42.8} & \textbf{42.1} & \textbf{79.6} & \textbf{78.7} & \textbf{55.6} & \textbf{54.8} \\
    2 & 2 & 8.6 & 6.6 & 62.4 & 55.4 & 35.5 & 35.3 & 66.2 & 66.2 & 46.2 & 46.0 \\
      & 3 & 17.4 & 15.5 & 66.9 & 61.0 & 38.9 & 38.3 & 72.5 & 71.9 & 50.6 & 50.0 \\\hline
      & 1 & 7.6 & 5.0 & 71.8 & 70.1 & \textbf{47.6} & \textbf{47.1 }& \textbf{88.7} & \textbf{88.3} & \textbf{61.9} & \textbf{61.4} \\
    64& 2 & 7.6 & 5.0 & 82.4 & \textbf{83.1} & 47.3 & 46.7 & 88.0 & 87.5 & 61.5 & 60.8 \\
      & 3 & 7.6 & 5.0 & \textbf{83.8} & 79.0 & 47.3 & 46.6 & 88.1 & 87.5 & 61.5 & 60.8 \\
    \bottomrule
\end{tabular}
  }
\caption{$t_x$-annotation-similarity for all models trained on all formal languages. }
\label{tab:t-annotation-similarity-formal}
\end{table*}

\begin{figure}
  \includegraphics[width=\linewidth]{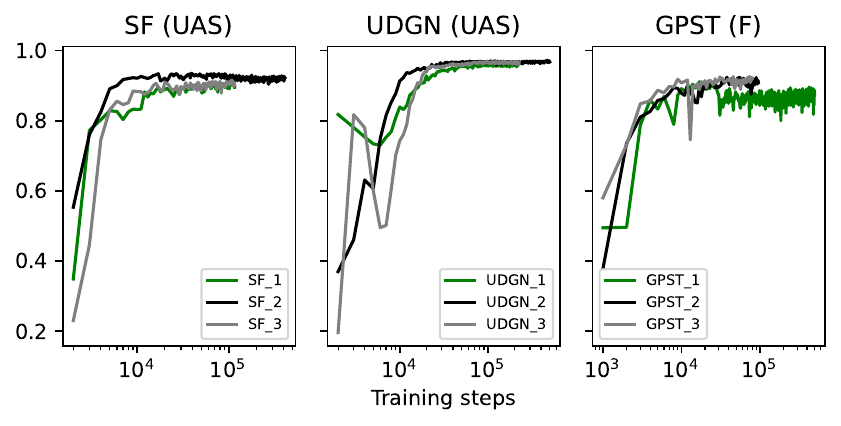}
  \caption{$t_x$-evolution for English}
  \label{fig:t-evo-english}
\end{figure}

\paragraph{$t_x$-learning-evolution} 
Figure~\ref{fig:t-evo-english} displays for all English models the similarity of $t_x$ induced from adjacent training checkpoints on a held-out test set. 
Checkpoints are evaluated every 1000 training steps. 
The first 20K training steps show the biggest changes in induced $t_x$, and after 50K steps, the $t_x$ have a high similarity - at least 87 points bracketing F score for \gpst, 89 points UAS for \sfrmr and 95 points UAS for \udgn. 
These scores stay relatively stable at high values, without reaching a point where trees do not change anymore when training further. 
This behavior can be attributed to different factors, such as random data presentation order during training, and the probabilistic nature of $H$ in \sfrmr and \udgn. 

For \gpst, a certain amount of volatility 
can be explained by the more complex loss function. 
Here, we observe that while the structural loss component is saturated relatively quickly, language modeling loss is generally more volatile even in later stages of training. 
This is also reflected in the $t_x$-evolution for formal languages (App., Table~\ref{tab:evolution-laststeps-formal}), with the difference that $t_x$ induced from \sfrmr models shows relatively low values. 
This is expected based on the fact that \sfrmr induces relatively unstable $t_x$ for the formal languages in general (see above) We also evaluated $t_x$-consistency, $t_x$-triviality, $t_x$-annotation-similarity and $t_x$-evolution for all other models and languages along all training runs, and generally find that these values stabilize at some point during training.

\begin{table*}\centering

\scalebox{.75}{
\begin{tabular}{l|rr|rrr|rrr|rrr}
\toprule
 & \sfrmr & \udgn  & \multicolumn{9}{c}{\gpst} \\
& &   & \multicolumn{3}{c}{left-factorized bin.} & \multicolumn{3}{c}{right-factorized bin.} & \multicolumn{3}{c}{no bin.} \\
& UAS & UAS & P & R & F  & P & R & F  & P & R & F \\
\midrule
1 & 23.2 & 19.5 & 42.3 & 40.3 & 41.3 & 40.6 & 38.6 & 39.6 & 31.1 & 52.8 & 38.7 \\
2 & \textbf{25.5} & \textbf{22.8} & \textbf{44.6} & 42.5 & \textbf{43.5} & 42.9 & 41.0 & 41.9 & 34.9 & \textbf{58.3} & 43.2 \\
3 & 14.4 & 19.5 & 43.6 & 41.7 & 42.6 & 38.9 & 37.3 & 38.0 & 33.1 & 55.4 & 41.1 \\
\bottomrule
\end{tabular}
}
\caption{$t_x$-annotation-similarity for English.}
\label{tab:t-annotation-similarity-english}
\end{table*}
\begin{table*}\centering
\scalebox{.75}{
\begin{tabular}{l|rr|rrr|rrr|rrr}
\toprule
 & \sfrmr & \udgn  & \multicolumn{9}{c}{\gpst} \\
& &   & \multicolumn{3}{c}{left-factorized bin.} & \multicolumn{3}{c}{right-factorized bin.} & \multicolumn{3}{c}{no bin.} \\
& UAS & UAS & P & R & F  & P & R & F  & P & R & F \\
\midrule
1 & 30.9 & 30.3 & 23.4 & 22.7 & 23.0 & \textbf{30.0} & 28.4 & \textbf{29.0}& 17.8 & \textbf{33.7} & 23.0 \\
2 & \textbf{33.1} & 27.3 & 18.2 & 17.8 & 17.9 & 23.8 & 22.9 & 23.3 & 14.9 & 27.8 & 19.1 \\
3 & 25.0 & \textbf{30.9} & 18.5 & 18.0 & 18.2 & 24.0 & 23.0 & 23.4 & 14.7 & 27.8 & 19.0 \\
\bottomrule
\end{tabular}
}
\caption{$t_x$-annotation-similarity for German.}
\label{tab:t-annotation-similarity-german}
\end{table*}

\paragraph{$t_x$-annotation-similarity} 
Here, we investigate how similar the induced $t_x$ are to parser outputs (for English and German), and to gold annotations (for formal languages). 
For the formal languages, we find that $t_x$ induced from \udgn have relatively high UAS towards the gold bracketing trees, except for trees in the Dyck-$1$ language (Table~\ref{tab:t-annotation-similarity-formal}). 
\sfrmr models, however, do not learn trees that are similar to gold annotations. 
\gpst models also learn trees that are similar to the gold constituency trees. 
Since \gpst models induce binary trees but the grammar that generates bracketing structures is not necessarily binary (see above), bracketing recall is much higher than precision. However, the relatively high recall %
 values show that in principle, \gpst induces constituents for input substrings defined as constituents by the annotations. 
This holds for both validation and generalization sets, and for all languages. 

For the Dyck-$k$ languages, for both \gpst and \udgn models, $t_x$-annotation-similarity is highest for Dyck-$64$.
For English, we find that $t_x$ from \sfrmr and \udgn have generally low similarity to parser outputs (Table \ref{tab:t-annotation-similarity-english}).
\gpst, on the other hand, induces constituent trees that have reasonably high precision, recall and F-score with trees obtained with a constituent parser. 
To control for the fact that trees induced from \gpst are binarized while trees obtained with the SuPar parser are not, we also evaluate against left-factored and right-factored binarizations of the trees obtained from the parser. 
We find that there is a small tendency (between 2.6-2.7 points on each metric) towards left-factored binarizations. 
For German, we find that all induced $t_x$ have generally low similarity to parser outputs (Table \ref{tab:t-annotation-similarity-german}). 
On a smaller-scale qualititative investigation, we find that this is largely due to the treatment of non-projective dependencies. These non-context-free structures are frequent in German data, but \sfrmr and \gpst are both unable to represent them. 

\begin{table}\centering\scalebox{.75}{
\begin{tabular}{lrrrrr}
\toprule
lang & \alm & \tf & \sfrmr & \udgn & \gpst \\
\midrule
en & 73.6 & \textbf{76.3} & 73.5 & 69.8 & 72.4 \\
de & 95.8 & 96.0 & 93.4 & 87.2 & \textbf{96.4} \\
zh & 53.4 & 77.4 & 76.6 & 69.6 & \textbf{78.5} \\
\bottomrule
\end{tabular}
}
\caption{Performance on minimal pair benchmarks, by language.}
\label{tab:blimp-by-lang-aggregated}
\end{table}

\paragraph{Performance: English, German, Chinese} 
We evaluate the natural language models with respect to perplexity (for \gpst and \alm) and masked language modeling pseudo-perplexity \cite{salazar-etal-2020-masked}. 
We find that, for all models, these metrics gradually improve over the whole training run. 
The transformer encoder baseline \tf outperforms both \sfrmr and \udgn (App.~\ref{app:moreresults}). 
For minimal pair evaluation benchmarks, we display the performance aggregated by language in Table~\ref{tab:blimp-by-lang-aggregated}. 
On English, \udgn is the worst-performing model quite consistently across categories (Tab.~\ref{tab:blimp-by-cat}). 
\tf outperforms \gpst on 7 out of 12 categories, and within the categories, the best model is always \tf, \alm, or \gpst. 
In English, \sfrmr performance patterns across categories are similar to \tf, which means that \sfrmr outperforms \gpst overall but \tf always scores higher than \sfrmr. These patterns are generally repeated for German (Tab.~\ref{tab:germanblimp-by-cat}) and Chinese (Tab.~\ref{tab:chinese-blimp-by-cat}). 

\begin{table}\centering\scalebox{.73}{
\begin{tabular}{lrrrrr}
\toprule
category & $alm$ & $tf^{l2r}$ & $sf_2^{l2r}$ & $udgn_2^{l2r}$ & $gpst_1$ \\
\midrule
anaphor agreement & \textbf{95.8} & 87.8 & 85.9 & 75.6 & 87.8 \\
argument structure & \textbf{75.5} & 72.7 & 69.7 & 64.2 & 73.7 \\
binding & \textbf{73.5} & 71.1 & 71.5 & 69.8 & \textbf{73.5} \\
control raising & 69.6 & \textbf{69.7} & 68.2 & 63.2 & 68.2 \\
det noun agreement & 88.5 & \textbf{93.8} & 91.2 & 87.0 & 83.0 \\
ellipsis & 68.3 & \textbf{80.7} & 75.8 & 75.8 & 59.2 \\
filler gap & 70.6 & \textbf{75.1} & 73.8 & 68.8 & 67.7 \\
irregular forms & 94.2 & \textbf{99.1} & 98.5 & 91.4 & 97.0 \\
island effects & 52.4 & \textbf{60.2} & 51.9 & 55.0 & 50.6 \\
npi licensing & 66.2 & \textbf{78.6} & 73.5 & 69.6 & 71.2 \\
quantifiers & \textbf{72.5} & 64.2 & 62.9 & 64.9 & 69.0 \\
subject-verb agreement & 83.4 & 85.6 & 84.2 & 73.8 & \textbf{87.6} \\\midrule
overall mean & 73.6 & \textbf{76.3} & 73.5 & 69.8 & 72.4 \\
\bottomrule
\end{tabular}
}
    \caption{Accuracy on BLiMP, aggregated by category}
    \label{tab:blimp-by-cat}
\end{table}

\begin{figure}\centering
  \includegraphics[width=\linewidth]{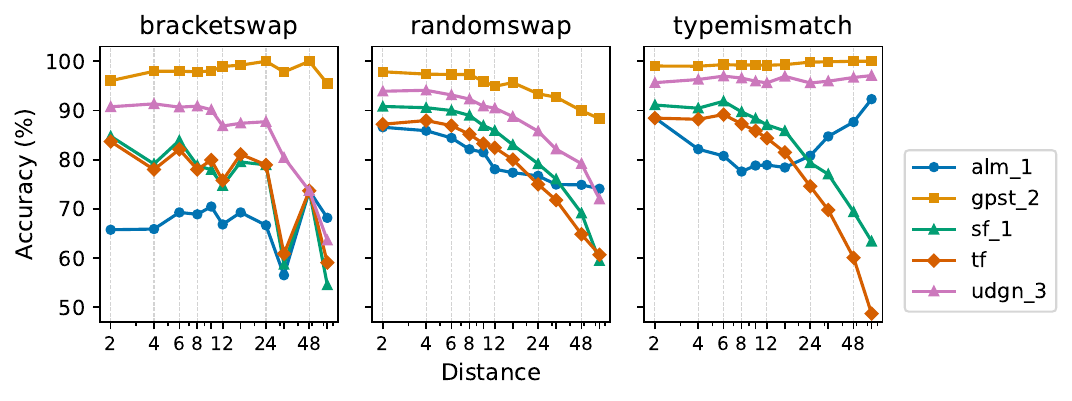}
  \caption{Performance on minimal pairs for Dyck-$u$, by distance between the brackets. The smallest distance, 2, refers to the case where the brackets are adjacent.}
  \label{fig:mpdyck-u-model-comparison}
\end{figure}

\paragraph{Performance on Dyck languages} 
We evaluate the formal language model performance using minimal pairs. %
We find that the performance differences between models of the same architectures trained on the same language is always within 3 points of accuracy, and therefore we only display the best-performing model per architecture. 
On validation sets, we observe accuracies of close to $100\%$ for all languages and models. 

Performance on generalization sets with different distances between the perturbed brackets is displayed in Figure~\ref{fig:mpdyck-u-model-comparison}. 
We observe that, with almost no exceptions and for all perturbations, \sfrmr has the lowest performance (slightly better than transformer encoder and decoder baselines). The transformer decoder baseline \alm performs worst, except for long-distance dependencies.
\udgn outperforms the transformer encoder baseline by 8.8 points accuracy (on the bracketswap subtask), and
\gpst receives the highest performance across all settings. 
In addition, \gpst shows the lowest performance drop when the ungrammatical example contains a long-distance dependency. 
This evaluation shows that the Dyck-$u$ languages are a useful benchmark for comparing different structure inducing models since they focus specifically on syntactic generalization capabilities.
Results for the other formal languages are displayed in App.~\ref{app:moreresults}.

\paragraph{Scaling behavior and training dynamics}

\begin{table}\centering
  \scalebox{.75}{
  \begin{tabular}{l|rrrr}
\toprule
        & \tf & \sfrmr & \udgn & \gpst \\
\midrule
Dyck-64 & \textbf{7} & 38 & 8 & 27 \\
English & 10 & 14 & \textbf{6} & 28 \\
\bottomrule
\end{tabular}
  }
  \caption{Training time in minutes for 1000 steps}
  \label{tab:train-efficiency}
\end{table}

We find that training speed (Table \ref{tab:train-efficiency}) depends on model size, model architecture, as well as maximum sequence length. 
The two settings differ in that for Dyck-64 small models are trained but the sequence length is longer than for English. 
\udgn is generally the fastest \model in training, and \sfrmr is very slow on Dyck-64. 
This is because the computation of possible heads for \sfrmr has a space complexity of $n^3$ for a maximum sequence length $n$, while \udgn and \gpst only require $n^2$. 
This requires \sfrmr formal language models to be trained with lower batch sizes (and more gradient accumulation steps) on the same hardware. 
\gpst is comparatively slow on both datasets. 

In terms of training dynamics, we observe for all models that the relevant loss functions are continually declining over the course of training, and are still slowly declining at the end of the allocated training time. 
Together with the fact that $t_x$ change little near the end of training, we interpret this such that the training is stable. Even longer training could potentially yield small improvements. 

\paragraph{Additional experiments} In App.~\ref{app:moreresults}, we present additional results. We evaluate the original \sfrmr implementation, where the parser module is applied at the embedding layer \cite{shen-etal-2021-structformer}. We explain the low $t_x$ induction performance of \sfrmr on  formal languages, and we perform a direct comparison between constituency trees induced from \gpst and \sfrmr by evaluating both via bracketing F score.

\section{Summary and Conclusion}\label{sec:conclusion}

\paragraph{Contributions} 
We evaluate Structure-Inducing Language Models (\model) on three natural languages (English, German, and Chinese) as well as formal bracketing expressions (Dyck languages). 
We introduce two datasets for formal languages. 
First, inspired by underspecification in English number agreement, we introduce the Dyck-$u$ language. 
In Dyck-$u$, bracket types can be underspecified, which forces models to keep track explicitly of the hierarchical structure underlying the bracketing. 
Second, to connect the recognition task in formal languages with the probabilistic nature of masked and autoregressive language modeling, we introduce a minimal pair benchmark for Dyck languages, with controlled perturbations to test the model abilities to represent bracketing structure with different bracketing distances. 
Evaluation is concerned with isolating the models abilities to learn and generalize hierarchical structures (via formal languages), as well as their capabilities when trained on different natural languages. 
Evaluation focuses on the properties of induced syntactic representations ($t_x$), as well as performance on minimal pair benchmarks. 

Results show that there are nontrivial differences between training \model s on different kinds of datasets.
Most crucially, we find that all tested \model\ architectures induce syntactic representations that change across different training runs. 
None of the three \model\ architectures stands out across evaluation settings. 
The \gpst architecture is outperformed by a transformer encoder baseline on English minimal pairs. It performs well on German and Chinese minimal pair benchmarks, and long-distance dependencies in formal languages.  

\paragraph{Induced syntactic representations} We find that $t_x$ are generally manifested in earlier stages of training and after that change relatively little. 
Except for some exceptions, $t_x$ are not identical to trivial baseline trees. 
We find that none of the \model\ architecture induces syntactic representations that are perfectly consistent over several training runs. 
Moreover, \sfrmr models fail to induce meaningful nontrivial $t_x$ on all formal languages, and both \sfrmr and \udgn fail to induce consistent $t_x$ for the length generalization datasets in the formal languages. 
On the formal languages, \sfrmr does not induce dependency distributions that match the bracketing structure of the underlying data distribution. 
Both \udgn and \gpst induce syntactic representations that are similar to the underlying gold distributions, with the highest similarity for Dyck-64. 
This suggests that, for evaluations on formal languages, very small vocabularies potentially harm induction capabilities.  
On English, we find that both \sfrmr and \udgn induce $t_x$ that are dissimilar to parsed dependency trees. This has several reasons. 
First, both put significant weight in $H$ on tokens in the same multi-token word, which means that other words are receiving much lower weights in $H$. 
\gpst, on the other hand, is already designed such that multi-token words form a constituent. 
Second, the dependency distribution $H$ is probabilistic, and it is not guaranteed that selecting the most likely head for each token leads to a dependency tree structure\footnote{For both models, less than $7\%$ of the dependency graphs obtained from $t_x$ are fully connected trees}. 
\gpst, on the other hand, induces binary constituent trees that have a reasonably high similarity with parser outputs, and do not show strong left- or right-branching biases. 
We hypothesize that there is a connection between $t_x$-consistency and $t_x$-evolution: Self-supervised training fails to induce syntactic representations that are perfectly consistent between model instances, and representations still change considerably even after relatively long training, and with little changes on validation loss and perplexity. 
This suggests that - to different degrees, depending on the model - $t_x$ contains subsequences for which stable syntactic representations are hard to find. 

\paragraph{Performance} In terms of model performance, we find that a transformer encoder baseline \tf outperforms all other models in terms of English language modeling (pseudo-)perplexity. 
On the English minimal pair benchmark BLiMP, \tf outperforms all \model s, and on non-English minimal pair benchmarks, \gpst performs best. 
\citet{warstadt-etal-2023-findings} have shown that this range of performance of BLiMP correlates with an aggregated performance on a subset of GLUE and SuperGLUE \cite{wang-etal-2018-glue,wang-etal-2019-superglue} of .7 or higher. Therefore the models we train here can be expected to perform well on other downstream tasks.
On formal language minimal pair evaluation, \gpst clearly outperforms the other models. 
This results in a mixed picture where all models have certain strengths and weaknesses. 
However, \gpst is the only model that performs relatively consistently across our evaluations: 
(i) \gpst induces non-trivial trees,
(ii) it generalizes well to longer sequences and long-distance dependencies on formal languages, 
(iii) it induces trees that are reasonably similar to parsed and gold constituency trees, and 
(iv) it outperforms a transformer baseline on some linguistic phenomena in English minimal pairs, and all other models in general performance on German and Chinese minimal pairs. 
\sfrmr and \udgn, on the other hand, have clear weaknesses in terms of induced $t_x$ and performance.

\section{Open Issues and Future Directions}\label{sec:openissues}

\paragraph{Robustness}
All evaluated \model s have weaknesses with respect to the induced structures. 
In order for the induced representations to be useful, one would expect that these representations are stable when training models repeatedly on the same data. 
However, this is clearly not the case. 
This has several implications: 
\citet{williams-etal-2018-latent}'s finding that $t_x$-consistency has to be taken into account when developing \model s is confirmed for the more recent architectures trained here.
Especially for formal languages, it is concerning that no stable structures can be induced. 
For natural languages, additional work is needed to show which linguistic phenomena lead to stable structures, and for which phenomena this is more difficult. 
If high $t_x$-consistency cannot be obtained for models trained on natural languages, the reason can be a weakness of the architecture or training process, or that estimating the underlying hierarchical sentence structure of the training data is simply an ambiguous and error-prone process.

\paragraph{Evaluation} 
Existing \model s have been applied to a number of diverse tasks  evaluating linguistic skills such as linguistic generalizations \cite[among others]{warstadt-etal-2020-blimp,hu-etal-2020-systematic}, unsupervised parsing, various traditional NLP benchmarks, and tasks related to synthetic data.
However, the majority of natural language evaluations focus on English, which has been shown to be insufficient with respect to many capabilities such as word and constituent order, tokenization and morphology, etc. 
English parsing evaluations are typically conducted on standard English PennTreebank splits, which has been shown to overestimate model performance in various settings \cite{coltekin-2020-verification,gorman-bedrick-2019-need}.
Moreover, those parsing evaluations sometimes involve finetuning the model on the treebank text data, which can skew the syntax induction capabilities of the model \cite{hu-etal-2024-generative,sinha-etal-2021-masked}. 

\paragraph{Improving Scalability}
GPST performs most consistently across our evaluations, however it is also the slowest to train. 
To build large-scale \model s, training efficiency is hugely important. 
As a consequence, future efforts also need to explore the efficiency gains and effects of accelerating the \model\ training pipeline with tools such as low floating point precision training, scaling parts of the \model\ architecture in size, etc.

\bibliographystyle{acl_natbib}
\bibliography{tacl2021,custom,anthology}

\onecolumn

\appendix

\section{Model details\label{app:modeldims}}

On English and Dyck-$u$, we have experimented with batch sizes between 128 and 8192, and learning rates between $1*10^{-5}$ and $5*10^{-4}$. 
For batch sizes, we observed no significant differences after early training stages. 
For learning rates, we observed that rates larger than $5*10^{-5}$ led to unstable behavior in later training stages on both languages - consistenly across models and baselines. 
Furthermore, we have experimented with learning rate warmup on English \model s but have not observed an improvement in performance. 
Table~\ref{tab:pretrain-hps} displays training hyperparameters. Since we did not find significant performance differences between architectures wrt.~these hyperparameters, we chose the same values for all models. 
We have fixed total parameter count, model dimensions, and vocabulary sizes (Table~\ref{tab:modeldims}) to ensure comparability between models and languages as much as possible. 

\begin{table}\centering
  \begin{center}
      \begin{tabular}{lr}
        \toprule
          Batch size & 1024 \\
          Initial Learning rate & $5*10^{-5}$ \\
          LR Scheduler & linear \\
          Masking rate & $15\%$ \\
          Vocabulary size (en, de, zh) & 10000 \\
          Vocubulary size (Dyck) & $2*k + |\{BOS,EOS,PAD,MASK\}|$ \\
          \bottomrule
      \end{tabular}
      \caption{Pre-training hyperparameters}
      \label{tab:pretrain-hps}
      \end{center}
  \end{table}
  
\begin{table}\centering
\begin{tabular}{lrr}
\toprule
\sfrmr & Dyck & En \\
\midrule
parser convolution size & 9 & 9 \\
parser layers & & 4 \\
$l_front$ & 3 & 3 \\
$l_back$ & 3 & 3 \\
attention heads & 8 & 8 \\
hidden size, embedding size transformer & 128 & 256 \\
\bottomrule
\end{tabular}
\begin{tabular}{lrr}
\toprule
\udgn & Dyck & En \\
\midrule
embedding size & 128 & \\
DGN layers & 4 & 4 \\
LSTM layers & 3 & 3 \\
DGN heads & 8 & 8 \\
DGN head size & 32 & 32 \\
\bottomrule
\end{tabular}
\begin{tabular}{lrr}
\toprule
\gpst & Dyck & En \\
\midrule
all hidden sizes for $TF$ blocks & 96 & 256 \\
all number of attention heads for $TF$ blocks & 4 & 8 \\
inner dimensions FFN blocks & 96 & 512 \\
$TF_{action}$ layers & 3 & 3\\
$TF_{ntp}$ layers & 3 & 5\\
parser layers & 3 & 3\\
\bottomrule
\end{tabular}
\caption{Model dimensions, if diverging from the respective original implementations.}
\label{tab:modeldims}
\end{table}

\newpage 
\section{Related work}\label{app:relatedwork}

A wide range of works connects unsupervised parsing and language modeling tasks.%
In earlier work, non-neural models have been proposed %
  \cite{charniak-2001-immediate,chelba-jelinek-2000-structured,klein-manning-2004-corpus}. 
Related architectures exist that do not fall under our definition for \model s. 
These include access to syntactic annotations at training time \cite{yoshida-etal-2024-tree}, semi-supervised approaches to syntactic representations \cite{corro-titov-2019-differentiable}, and information about other linguistic features such as grammaticality models \cite{cao-etal-2020-unsupervised}
 or POS-tags \cite{han-etal-2020-survey,grave-elhadad-2015-convex}.
Neural approaches can be classified between 
generative models (modeling the joint probability $p(x,t_x)$) and
discriminative approaches (modeling the conditional $p(t_x|x)$)
\cite{tu-etal-2021-unsupervised,han-etal-2020-survey}.
While no comprehensive taxonomy has been developed, related works can be partitioned into several broad categories depending on %
 the central backbone of the neural architectures used. Here, we highlight related approaches that are not mentioned in Section~\ref{sec:ltlms}.

\paragraph{RNN-based architectures}
Unsupervised constituency and dependency parsing has been approached using RNNs, LSTMs, and other recurrent structures. %
In addition to approaches mentioned above, proposals include studies investigating
reinforcement learning \cite{yogatama-etal-2017-learning},
inside-outside score computation \cite{le-zuidema-2015-unsupervised}, 
and the relation between hierarchical sequence structure and temporal information flow in recurrent models \cite{koutnik-etal-2014-clockwork,shen-etal-2019-ordered,schmidhuber-1991-neural,zhang-2020-latent}.
The majority of these works yield positive results. However, \citet{shi-etal-2018-tree} investigate the effect of injecting trivial trees in different Tree-LSTMs, and show that this injection leads to outperforming models with explicit syntactic information. 

\paragraph{Transformer-based architectures}
Beyond the architectures mentioned above, 
\citet{gu-etal-2022-phrase} extend the StructFormer and TreeTransformer \cite{wang-etal-2019-tree} models for constituency parsing. In particular, they use a parsing loss for tree distance, and find that the StructFormer outperforms the Tree-Transformer on several unsupervised constituency parsing metrics. 

\paragraph{Parsing-inspired architectures}
Inspired by traditional parsing algorithms, 
neural approaches to \citet{klein-manning-2004-corpus}'s dependency model with valence (DMV) have been proposed \cite{han-etal-2017-dependency,han-etal-2019-enhancing,jiang-etal-2017-combining,yang-etal-2020-second} for dependency parsing, 
as well as neural induction of probabilistic context-free grammars \cite{jin-etal-2021-character-based,kim-etal-2019-compound,zhu-etal-2020-return}.

\paragraph{\model\ evaluation}
\citet{li-etal-2020-empirical} compare a variety of unsupervised constituency tree parsing methods and find, on English and Japanese, that in terms of constituent tree induction, more recent models perform similarly in terms of induced tree F-Scores with gold tree, and recent neural models do not outperform older statistical models such as CCM \cite{klein-manning-2002-generative}. This suggests that some aspects of recent performance improvements are not due to better capturing of syntactic behavior, but due to data, scalability, and model sizes.

\section{Additional Results}\label{app:moreresults}

\paragraph{Results for classic StructFormer}
The experiments in the main paper featured \sfrmr variants for which the parsing layers were applied only after several sequential transformer encoder layers. 
The original definition by \cite{shen-etal-2021-structformer}, on the other hand, puts the parser layers directly after the embedding layers (such that parser layer inputs are not influenced by context). 
We have trained these \textit{classic} variants from \citet{shen-etal-2021-structformer} on both English (for 100K steps) and Dyck-$u$ data (for 45K steps), and found that the induced $t_x$ are different than those from our main experiment. 
In particular, English $t_x$ are much more similar to dependency trees in which the first (BOS) or last (EOS) token is the head. 
Depending on whether they are oriented towards, the similarity between $t_x$ from three different models ranges from 36 UAS to 66 UAS. 
For Dyck-$u$, classic \sfrmr show the same property as the \sfrmr in the main paper: Induced $t_x$ show uniform head distributions, resulting in wildly dissimilar trees for any model pair. 

\paragraph{Uniform head distributions in \structformer}\label{app:sf-uniformheads}

\structformer produces syntactic representations with low $t_x$-consistency for Dyck because the parsing module often does not put significant weight on dependency edges to other tokens. 
Both parsing modules in UDGN and StructFormer are designed such that initially, they put non-zero edge probabilities on the currently processed token, and in a subsequent step these edges are zeroed out to make sure that the parser module does not predict edges from a token to itself. 
For \sfrmr trained on Dyck languages, the parser module regularly induces large probabilities of edges towards itself (i.e., the main diagonal in $H$). 
Because values outside the main diagonal are not renormalized, all other values in the edge distribution matrix $H$ remain small. 
For instance, for the \udgn[1] model trained on Dyck-$u$, more than half of the most likely induced edges per token receive a probability $p(i,j)>.6$, whereas for the \sfrmr[1] model, $90\%$ of the edge probabilities are smaller than $.17$. 
Effectively, this means that the edge distribution matrices $H$ for StructFormer are so close to uniform distributions that no $t_x$ can be reliably induced.

\paragraph{StructFormer evaluation via constituency trees}
We have evaluated \sfrmr trained on natural languages also with respect to bracketing F scores when deriving constituency trees from the distance metric induced by the model, as described in \cite[Sec.~3]{shen-etal-2021-structformer}. 
The motivation is to put the metric difference -- UAS for \sfrmr and \udgn, F score for \gpst -- into perspective. 
The results are displayed in Tab.~\ref{tab:fscore-sfrmr}. We find that there is always a reasonable degree, and sometimes a high degree, of similarity between constituency trees induced from different training runs on each language. 
However, the scores for $t_x$-consistency in Tab.~\ref{tab:fscore-sfrmr} are lower than those for \gpst in Fig.~\ref{fig:t-consistency}, when taking the per-language mean. 
We also find that trees from \sfrmr and \gpst induced on the same data are reasonably similar. 

\begin{table}\centering
\begin{tabular}{l|rr}
  \toprule 
  \midrule
en  & 1 & 2 \\
3 & 52 & 72 \\
2 & 49 & \\
  \bottomrule
\end{tabular}
\begin{tabular}{l|rr}
  \toprule 
  \midrule
de  & 1 & 2 \\
3 & 86 & 74 \\
2 & 72 & \\
  \bottomrule
\end{tabular}
\begin{tabular}{l|rr}
  \toprule 
  \midrule
zh  & 1 & 2 \\
3 & 62 & 62 \\
2 & 72 & \\
  \bottomrule
\end{tabular}
\begin{tabular}{lrrr}
\toprule
  & en & de & zh \\
\midrule
  \sfrmr vs. \gpst & 43 & 52 & 50 \\
\bottomrule
\end{tabular}
\caption{$t_x$-consistency in constituency F scores for \sfrmr (left). F score for constituency trees of most-trained \sfrmr and \gpst for each language.}
\label{tab:fscore-sfrmr}
\end{table}

\begin{figure}[t]
  \centering
  \begin{minipage}[b]{0.3\textwidth}
    \centering
    \includegraphics[width=\linewidth]{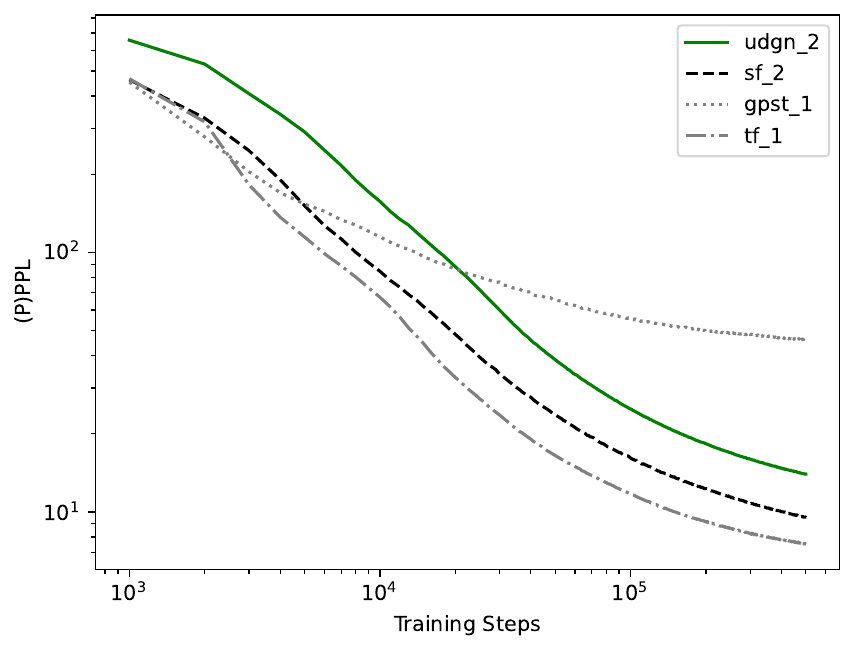}
    \caption{Test set Perplexity and Pseudo-Perplexity at checkpoints during training.}
    \label{fig:babylm-ppl}
  \end{minipage}
  \hfill
  \begin{minipage}[b]{0.58\textwidth}\centering
  \scalebox{.65}{
\begin{tabular}{ll|rrrrrr}
\toprule
& & \multicolumn{2}{c}{\sfrmr} & \multicolumn{2}{c}{\udgn} & \multicolumn{2}{c}{\gpst} \\
lang & model & val & gen & val & gen & val & gen\\
\midrule
     & 1 & 58.1 & 55.5 & 100.0 & 100.0 & 90.1 & 89.0 \\
1    & 2 & 49.2 & 45.7 & 100.0 & 99.6  & 93.7 & 86.3 \\
     & 3 & 70.3 & 68.7 & 100.0 & 99.9  & 92.2 & 88.5 \\
     & 1 & 59.8 & 61.0 & 90.4 &  86.0  & 89.6 & 95.5 \\
2    & 2 & 49.2 & 49.4 & 90.4 &  85.8  & 88.2 & 79.9 \\
     & 3 & 60.1 & 57.7 & 85.2 &  79.7  & 93.5 & 91.7 \\
     & 1 & 86.8 & 82.2 & 94.2 &  93.0  & 94.5 & 94.0 \\
64   & 2 & 96.6 & 87.7 & 95.7 &  94.4  & 92.9 & 92.6 \\
     & 3 & 78.6 & 75.9 & 95.2 &  94.1  & 92.6 & 92.0 \\
     & 1 & 77.9 & 70.7 & 98.1 &  92.9  & 85.0 & 90.3 \\
$u$  & 2 & 63.9 & 56.7 & 94.1 &  88.4  & 92.8 & 92.8 \\
     & 3 & 67.1 & 63.3 & 91.7 &  79.3  & 91.6 & 90.3 \\
\bottomrule
\end{tabular}
  }
    \caption{Mean UAS (\sfrmr, \udgn) and F score (\gpst) to the previous checkpoint across the last 10 checkpoints of training each model for formal languages.}
    \label{tab:evolution-laststeps-formal}
\end{minipage}
\end{figure}

\begin{figure}%
  \centering
  \includegraphics[trim=0.0cm 0.0cm 4.0cm .0cm, height=2.2cm]{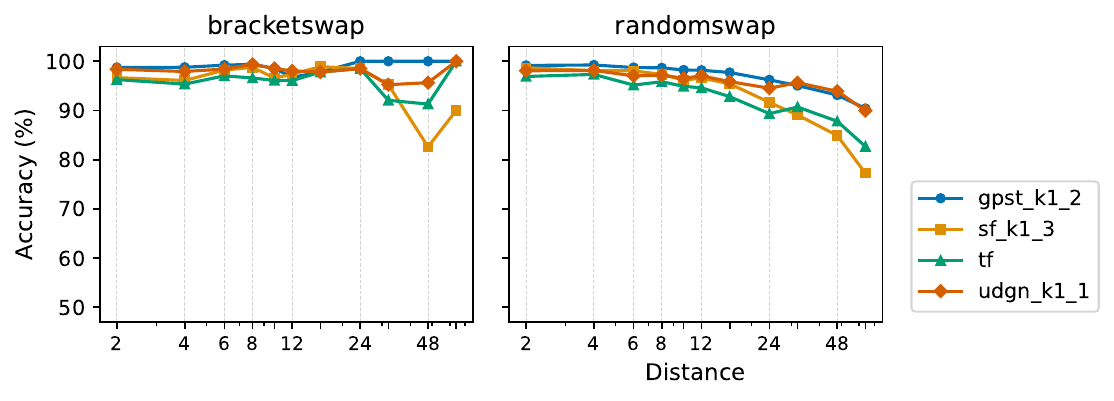}
  \includegraphics[trim=0.0cm 0.0cm 4.0cm .0cm, height=2.2cm]{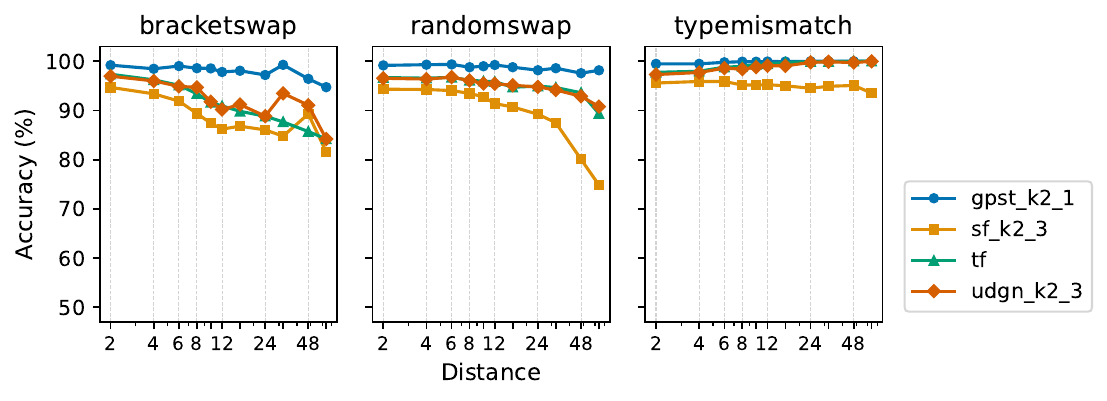}
  \includegraphics[trim=0.0cm 0.0cm 0.0cm .0cm, height=2.2cm]{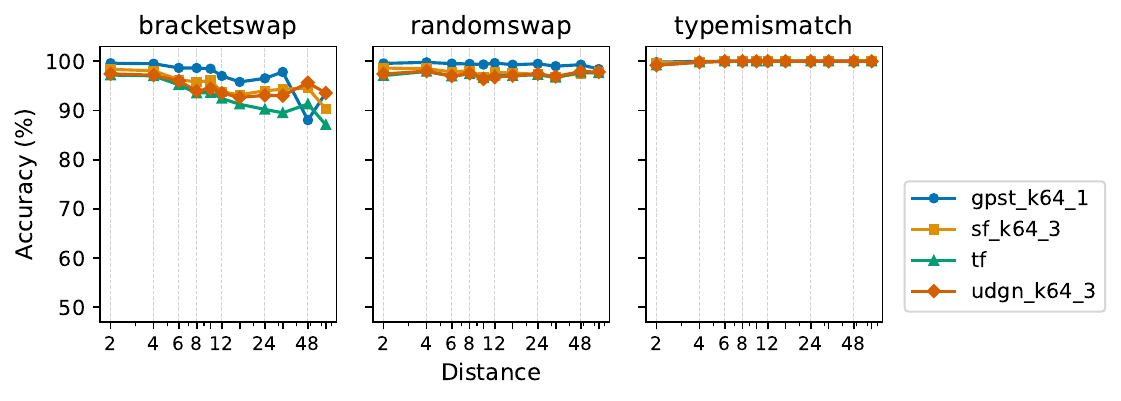}
  \caption{Results for minimal pair evaluation for Dyck generalization sets. From left to right: $k=1$, $k=2$, $k=64$.}
  \label{fig:blimpdyck}
\end{figure}

\paragraph{Other results}
Figure \ref{fig:babylm-ppl} displays the development over course of training for the best-performing models.
It shows that for all models, perplexity gradually improves, and still slightly improves near the end of training. 
\gpst is evaluated using perplexity (with only the left context available) and \sfrmr and \udgn have bidirectional conetxt available. This is the main reason that the absolute scores for \gpst are higher. 
Figure~\ref{fig:blimpdyck} displays the minimal pair evaluation on Dyck-1, Dyck-2, and Dyck-64. Because Dyck-1 has only one bracket type, the \texttt{typemismatch} subtask is not available.
Table~\ref{tab:blimp-full} displays performance on all models on all phenomena of English BLiMP, and 
Tables \ref{tab:blimp-by-cat}, \ref{tab:germanblimp-by-cat} and \ref{tab:chinese-blimp-by-cat} display performance by category per language.

\begin{table}\scalebox{.65}{
            \begin{tabular}{llrrrrrrrr|}
        \toprule
        \sfrmr & & \multicolumn{2}{c}{first} & \multicolumn{2}{c}{last} & \multicolumn{2}{c}{prev} & \multicolumn{2}{c}{next}\\
        lang & M & val & gen & val & gen & val & gen & val & gen \\
        \midrule
            & 1 & 3.1 & 0.0 & 7.0 & 2.7 & 6.8 & 5.0 & 12.0  & 4.7\\
        1   & 2 & 25.2 & 15.9 & 0.0 & 0.0 & 6.2 & 4.4 & 3.1 & 2.4  \\ 
            & 3 & 0.0 & 0.0 & 55.8 & 58.0 & 2.7 & 2.0 & 4.2 & 3.2 \\\hline
            & 1 & 0.0 & 0.0 & 0.0 & 0.0 & 6.4 & 5.2 & 4.6 & 3.7  \\ 
        2   & 2 & 6.2 & 4.3 & 0.0 & 0.0 & 8.9 & 6.7 & 8.5 & 6.5  \\ 
            & 3 & 0.6 & 0.1 & 1.2 & 1.0 & 15.1 & 13.6 & 16.1 & 14.3 \\\hline 
            & 1 & 90.7 & 57.3 & 9.3 & 5.9 &  3.2 & 2.3 & 0.1 & 0.7 \\ 
        64  & 2 & 91.2 & 57.0 & 8.8 & 6.1 & 3.6  & 2.6 & 0.3 & 0.5 \\
            & 3 & 77.9 & 48.3 & 22.1 & 14.9 & 3.0 & 2.1 & 1.3 & 1.1 \\\hline 
            & 1 & 2.4 &  1.5 & 26.8 & 16.2 & 6.2 &  8.1 & 10.8 & 11.2  \\ 
        $u$ & 2 & 11.9 &  6.4 & 0.0 & 0.0 & 4.2 & 2.7 & 3.0 & 2.3 \\ 
            & 3 & 6.5 &  3.8 & 0.0 & 0.0 & 7.6 & 4.9 & 6.1 & 4.3\\ 
        \bottomrule
        \end{tabular}
    \begin{tabular}{rrrrrrrr|}
        \toprule
         \multicolumn{2}{c}{\udgn first} & \multicolumn{2}{c}{last} & \multicolumn{2}{c}{prev} & \multicolumn{2}{c}{next}\\
        val & gen & val & gen & val & gen & val & gen \\
        \midrule
        50.0 & 50.0 & 50.0 & 50.0 & 0.0 & 0.0 & 0.0  & 0.0 \\
        50.0 & 50.0 & 50.0 & 50.0 & 0.0 & 0.0 & 0.0  & 0.0 \\
        50.0 & 50.0 & 50.0 & 50.0 & 0.0 & 0.0 & 0.0  & 0.0 \\\hline
        3.7 & 2.7 & 4.4 & 3.4 & 18.7 & 17.1 & 16.8 & 15.7 \\
        3.9 & 2.9 & 3.8 & 2.8 & 14.5 & 12.6 & 15.9 & 14.4 \\ 
        3.7 & 2.7 & 6.7 & 6.4 & 19.8 & 18.3 & 14.3 & 12.8 \\\hline 
        0.0 & 0.0 & 0.4 & 0.2 & 34.9 & 35.0 & 24.7 & 23.5 \\
        0.0 & 0.0 & 0.1 & 0.1 & 24.8 & 24.8 & 32.4 & 32.4 \\ 
        3.6 & 2.3 & 2.9 & 1.8 & 24.8 & 24.5 & 29.5 & 30.8 \\\hline 
        0.0 & 0.0 & 47.4 & 40.8 & 2.8 & 2.4 & 29.5 & 28.6 \\ 
        13.8 & 13.0 & 35.1& 33.9 & 3.0 & 2.3 & 27.8 & 24.9  \\ 
        19.0 & 25.4 & 4.6& 5.3  & 18.6 & 15.6 & 23.2 & 18.1  \\  
        \bottomrule
        \end{tabular}
            \begin{tabular}{rrrr}
        \toprule
      \multicolumn{2}{c}{\gpst left} & \multicolumn{2}{c}{right} \\
        val & gen & val & gen \\
        \midrule
             19.3 & 17.8 & 19.3 & 17.8 \\
             22.9 & 21.0 & 22.9 & 21.0 \\
             19.0 & 17.6 & 19.0 & 17.6 \\\hline
             13.5 & 12.6 & 13.5 & 12.6 \\
             17.4 & 16.0 & 17.4 & 16.0 \\
             16.2 & 15.0 & 16.2 & 15.0 \\\hline
             11.5 & 10.5 & 11.5 & 10.5 \\
             14.6 & 13.1 & 14.6 & 13.1 \\
             13.1 & 11.9 & 13.1 & 11.9 \\\hline
             8.1 & 7.9 & 8.1 & 7.9 \\
             11.9 & 11.1 & 11.9 & 11.1 \\
             14.5 & 13.3 & 14.5 & 13.3 \\
        \bottomrule
        \end{tabular}
        }
        \caption{F Scores to left-and right-branching binary trees on validation and generalization splits for \gpst trained on formal languages}
\end{table}

\begin{table}\centering\scalebox{.75}{
\begin{tabular}{lrrrrr}
\toprule
phenomenon & alm 1 & roberta 1 & structformer 3 & udgn 2 & gpstfullscaleloss 1 \\
\midrule
clams long vp coord & 99.8 & 98.8 & 100.0 & 93.4 & 97.8 \\
clams obj rel across anim & 99.4 & 95.6 & 90.6 & 75.4 & 99.9 \\
clams obj rel within anim & 86.5 & 97.8 & 95.4 & 89.8 & 87.9 \\
clams prep anim & 89.6 & 94.0 & 90.2 & 85.0 & 98.4 \\
clams simple agrmt & 100.0 & 95.7 & 92.9 & 91.4 & 100.0 \\
clams subj rel & 100.0 & 96.0 & 90.9 & 85.2 & 100.0 \\
clams vp coord & 99.2 & 95.3 & 91.5 & 86.9 & 98.3 \\
mblimp sv hash & 89.4 & 93.1 & 91.8 & 81.6 & 88.2 \\
mblimp sv p & 98.3 & 97.3 & 97.0 & 96.2 & 97.2 \\
overall mean & 95.8 & 96.0 & 93.4 & 87.2 & 96.4 \\
\bottomrule
\end{tabular}
}
\caption{Accuracy on German minimal pairs}
\label{tab:germanblimp-by-cat}
\end{table}

\begin{table}\centering\scalebox{.75}{
\begin{tabular}{lrrrrr}
\toprule
category & alm 1 & roberta 1 & structformer 1 & udgn 2 & gpst 1 \\
\midrule
BA & 59.7 & 88.1 & 86.3 & 79.2 & 88.0 \\
anaphor & 41.3 & 47.9 & 48.9 & 45.0 & 45.7 \\
argument structure & 58.3 & 76.4 & 73.1 & 68.1 & 73.9 \\
control raising & 52.0 & 94.4 & 94.1 & 94.3 & 94.7 \\
ellipsis & 74.2 & 36.5 & 54.0 & 65.8 & 75.7 \\
fci licensing & 40.0 & 89.9 & 91.6 & 87.1 & 96.0 \\
nominal expression & 71.3 & 85.8 & 88.6 & 89.4 & 89.2 \\
npi licensing & 62.2 & 45.9 & 43.5 & 26.6 & 54.6 \\
passive & 17.7 & 69.0 & 63.6 & 48.3 & 56.1 \\
quantifiers & 71.2 & 70.8 & 68.3 & 61.3 & 86.9 \\
question & 63.1 & 78.9 & 79.8 & 67.1 & 85.2 \\
relativization topicalization & 53.3 & 93.0 & 90.4 & 92.8 & 89.2 \\
verb phrase & 45.4 & 90.0 & 87.9 & 82.5 & 82.7 \\
overall mean & 53.4 & 77.4 & 76.6 & 69.6 & 78.5 \\ 
\bottomrule
\end{tabular}
}
\caption{Accuracy on Chinese minimal pairs, aggregated by category}
\label{tab:chinese-blimp-by-cat}
\end{table}

\begin{table*}\centering\scalebox{.72}{
\begin{tabular}{ll|r|rr|rr|rr|r}
\toprule
phenomenon & category & $alm$ & $tf$ & $tf^{l2r}$ & $sf_2$ & $sf_2^{l2r}$ & $udgn_2$ & $udgn_2^{l2r}$ & $gpst_1$ \\
\midrule
adjunct island & island effects & \textbf{77.8} & 57.9 & 59.2 & 48.1 & 49.4 & 63.8 & 66.2 & 40.7 \\
anaphor gender agreement & anaphor agreement & \textbf{92.8} & 80.6 & 78.8 & 78.8 & 76.7 & 70.0 & 67.1 & 80.0 \\
anaphor number agreement & anaphor agreement & \textbf{98.8} & 97.0 & 96.8 & 95.0 & 95.1 & 85.2 & 84.0 & 95.6 \\
animate subject passive & argument structure & 68.1 & 63.2 & \textbf{75.7} & 61.8 & 71.4 & 59.2 & 64.1 & 67.6 \\
animate subject trans & argument structure & \textbf{84.6} & 83.8 & 79.1 & 80.8 & 72.4 & 74.2 & 64.7 & 82.9 \\
causative & argument structure & 67.0 & \textbf{78.4} & 69.5 & 74.1 & 65.6 & 67.3 & 59.9 & 62.6 \\
complex NP island & island effects & \textbf{41.7} & 36.8 & 38.3 & 27.9 & 27.8 & 32.7 & 36.6 & 36.2 \\
coordinate structure constraint complex left branch & island effects & 30.3 & 63.8 & \textbf{70.5} & 43.7 & 48.8 & 36.4 & 39.2 & 32.5 \\
coordinate structure constraint object extraction & island effects & 71.7 & 90.5 & \textbf{92.5} & 89.4 & 90.8 & 84.3 & 86.3 & 90.6 \\
determiner noun agreement 1 & det noun agreement & 95.9 & 94.6 & \textbf{98.6} & 93.0 & 96.9 & 84.2 & 93.7 & 88.7 \\
determiner noun agreement 2 & det noun agreement & 95.5 & \textbf{99.7} & 99.6 & 99.5 & 99.6 & 98.5 & 98.2 & 90.0 \\
determiner noun agreement irregular 1 & det noun agreement & 80.8 & 84.0 & \textbf{89.8} & 80.2 & 84.1 & 72.9 & 73.5 & 72.1 \\
determiner noun agreement irregular 2 & det noun agreement & 89.9 & \textbf{91.4} & 87.1 & 86.9 & 83.0 & 87.3 & 84.7 & 82.2 \\
determiner noun agreement with adj 2 & det noun agreement & 92.7 & 97.4 & 97.2 & 97.6 & \textbf{97.9} & 92.8 & 93.1 & 86.0 \\
determiner noun agreement with adj irregular 1 & det noun agreement & 76.7 & 81.5 & \textbf{91.0} & 81.4 & 90.0 & 80.6 & 82.2 & 78.1 \\
determiner noun agreement with adj irregular 2 & det noun agreement & 85.0 & \textbf{91.5} & 89.6 & 85.2 & 83.2 & 85.8 & 84.6 & 80.2 \\
determiner noun agreement with adjective 1 & det noun agreement & 91.3 & 93.3 & \textbf{97.5} & 92.4 & 95.0 & 79.6 & 86.0 & 86.5 \\
distractor agreement relational noun & subject-verb agreement & 75.4 & 85.3 & 86.4 & 77.6 & 79.4 & 64.1 & 64.6 & \textbf{88.8} \\
distractor agreement relative clause & subject-verb agreement & 63.6 & 68.3 & 67.0 & 68.4 & 66.3 & 60.5 & 59.4 & \textbf{73.7} \\
drop argument & argument structure & \textbf{71.5} & 57.5 & 65.7 & 56.1 & 65.3 & 55.4 & 63.3 & 71.1 \\
ellipsis n bar 1 & ellipsis & 69.9 & \textbf{86.5} & 85.1 & 85.6 & 85.3 & 79.1 & 77.9 & 65.7 \\
ellipsis n bar 2 & ellipsis & 66.7 & \textbf{79.1} & 76.2 & 74.6 & 66.2 & 76.9 & 73.7 & 52.8 \\
existential there object raising & control raising & 74.5 & 72.5 & \textbf{76.7} & 65.5 & 75.5 & 59.0 & 69.8 & 75.2 \\
existential there quantifiers 1 & quantifiers & \textbf{99.1} & 97.7 & 97.7 & 98.5 & 98.3 & 98.8 & 98.5 & 98.9 \\
existential there quantifiers 2 & quantifiers & \textbf{26.3} & 17.4 & 15.8 & 17.1 & 17.1 & 11.5 & 8.8 & 24.9 \\
existential there subject raising & control raising & 81.6 & 76.2 & \textbf{84.0} & 79.1 & 81.5 & 66.2 & 72.3 & 78.1 \\
expletive it object raising & control raising & 70.2 & 69.6 & \textbf{72.7} & 65.7 & 70.5 & 62.3 & 67.3 & 72.1 \\
inchoative & argument structure & 57.5 & \textbf{67.7} & 49.8 & 60.3 & 47.1 & 50.0 & 41.0 & 56.7 \\
intransitive & argument structure & \textbf{71.2} & 66.2 & 55.1 & 61.6 & 52.9 & 51.3 & 45.9 & \textbf{71.2} \\
irregular past participle adjectives & irregular forms & 90.8 & 98.3 & \textbf{99.4} & 92.5 & 98.4 & 79.4 & 91.5 & 97.9 \\
irregular past participle verbs & irregular forms & 97.5 & 92.4 & \textbf{98.8} & 93.2 & 98.5 & 89.8 & 91.3 & 96.2 \\
irregular plural subject verb agreement 1 & subject-verb agreement & \textbf{87.8} & 86.5 & 86.4 & 82.3 & 83.4 & 77.8 & 76.0 & 87.5 \\
irregular plural subject verb agreement 2 & subject-verb agreement & \textbf{91.6} & 87.0 & 88.7 & 89.7 & 90.5 & 80.6 & 80.8 & 91.5 \\
left branch island echo question & island effects & 37.0 & 30.0 & 34.4 & 24.8 & 30.6 & 26.9 & 32.3 & \textbf{47.0} \\
left branch island simple question & island effects & 45.8 & 89.5 & \textbf{92.9} & 83.5 & 89.1 & 78.3 & 80.6 & 63.4 \\
matrix question npi licensor present & npi licensing & 24.5 & 80.4 & \textbf{80.5} & 74.5 & 76.3 & 49.2 & 47.1 & 30.1 \\
npi present 1 & npi licensing & 56.5 & 62.0 & \textbf{68.4} & 46.6 & 52.9 & 42.2 & 46.2 & 54.8 \\
npi present 2 & npi licensing & \textbf{71.0} & 63.7 & 67.2 & 54.3 & 60.3 & 44.1 & 50.3 & 61.2 \\
only npi licensor present & npi licensing & 93.6 & \textbf{100.0} & \textbf{100.0} & 78.6 & 73.2 & 75.4 & 83.2 & 94.5 \\
only npi scope & npi licensing & 79.6 & 80.5 & 82.9 & 77.3 & 81.1 & 87.8 & \textbf{92.2} & 87.3 \\
passive 1 & argument structure & \textbf{89.1} & 71.2 & 84.0 & 69.2 & 83.2 & 65.8 & 77.5 & 87.8 \\
passive 2 & argument structure & 85.7 & 78.0 & \textbf{89.9} & 73.8 & 86.8 & 71.1 & 81.1 & 84.1 \\
principle A c command & binding & 66.0 & 58.7 & 53.6 & 61.1 & 61.1 & 57.5 & 56.8 & \textbf{67.9} \\
principle A case 1 & binding & \textbf{100.0} & \textbf{100.0} & \textbf{100.0} & \textbf{100.0} & \textbf{100.0} & \textbf{100.0} & 99.9 & 99.9 \\
principle A case 2 & binding & 89.2 & 95.5 & \textbf{97.1} & 96.6 & 96.9 & 94.9 & 93.9 & 91.5 \\
principle A domain 1 & binding & \textbf{99.2} & 96.3 & 93.7 & 98.0 & 95.9 & 97.0 & 95.2 & 98.4 \\
principle A domain 2 & binding & 62.8 & 65.7 & \textbf{66.8} & 61.8 & 61.5 & 61.5 & 62.9 & 61.5 \\
principle A domain 3 & binding & \textbf{59.3} & 46.2 & 56.9 & 46.3 & 55.3 & 48.6 & 57.7 & 47.3 \\
principle A reconstruction & binding & 38.3 & 20.6 & 29.7 & 21.3 & 29.9 & 13.0 & 22.1 & \textbf{47.9} \\
regular plural subject verb agreement 1 & subject-verb agreement & \textbf{95.5} & 90.9 & 94.5 & 89.8 & 93.6 & 80.8 & 83.4 & 94.8 \\
regular plural subject verb agreement 2 & subject-verb agreement & 86.2 & 92.4 & 90.6 & \textbf{93.2} & 92.3 & 76.7 & 78.8 & 89.4 \\
sentential negation npi licensor present & npi licensing & \textbf{100.0} & 84.0 & \textbf{100.0} & 93.2 & \textbf{100.0} & 87.4 & \textbf{100.0} & 99.0 \\
sentential negation npi scope & npi licensing & 38.0 & 46.9 & 51.1 & 65.9 & 70.7 & 68.5 & 68.5 & \textbf{71.2} \\
sentential subject island & island effects & 35.2 & 45.9 & 43.6 & 45.8 & 42.8 & 51.9 & \textbf{56.7} & 48.2 \\
superlative quantifiers 1 & quantifiers & \textbf{85.6} & 74.8 & 72.5 & 66.2 & 66.8 & 73.8 & 73.5 & 79.2 \\
superlative quantifiers 2 & quantifiers & 79.0 & 73.0 & 70.7 & 69.9 & 69.3 & \textbf{81.4} & 78.9 & 73.2 \\
tough vs raising 1 & control raising & 36.3 & 46.9 & 34.2 & 49.5 & 37.1 & \textbf{61.3} & 45.3 & 34.6 \\
tough vs raising 2 & control raising & \textbf{85.4} & 65.7 & 81.0 & 60.3 & 76.3 & 42.6 & 61.3 & 81.0 \\
transitive & argument structure & 84.8 & 73.1 & \textbf{85.3} & 72.5 & 82.2 & 71.0 & 80.1 & 79.5 \\
wh island & island effects & \textbf{79.3} & 48.6 & 50.3 & 39.1 & 35.8 & 40.2 & 42.4 & 46.3 \\
wh questions object gap & filler gap & 73.4 & 80.4 & \textbf{84.7} & 77.1 & 80.6 & 60.6 & 66.0 & 71.3 \\
wh questions subject gap & filler gap & \textbf{93.3} & 90.4 & 93.2 & 82.2 & 86.0 & 79.3 & 86.0 & 85.2 \\
wh questions subject gap long distance & filler gap & \textbf{94.9} & 84.7 & 84.2 & 88.5 & 89.0 & 92.4 & 93.3 & 86.0 \\
wh vs that no gap & filler gap & 99.2 & 98.3 & 98.9 & 98.7 & \textbf{99.3} & 97.0 & 97.5 & 98.4 \\
wh vs that no gap long distance & filler gap & \textbf{99.3} & 97.9 & 98.6 & 97.8 & 98.5 & 97.3 & 97.9 & 97.2 \\
wh vs that with gap & filler gap & 26.8 & \textbf{49.9} & 48.0 & 46.0 & 45.1 & 29.4 & 31.0 & 22.2 \\
wh vs that with gap long distance & filler gap & 7.2 & \textbf{21.5} & 18.0 & 19.5 & 18.0 & 10.2 & 9.7 & 13.3 \\\midrule
overall mean &   & 73.6 & 74.6 & \textbf{76.3} & 71.7 & 73.5 & 67.7 & 69.8 & 72.4 \\
\bottomrule
\end{tabular}
        }
        \caption{Performance on BLiMP}
        \label{tab:blimp-full}
\end{table*}

\end{document}